\documentclass{article}
\PassOptionsToPackage{numbers, compress}{natbib}
\usepackage[final]{neurips_2022}

\usepackage{microtype}
\usepackage{siunitx}
\usepackage{graphicx}
\usepackage{booktabs}
\usepackage{hyperref}

\usepackage{amsmath}
\usepackage{amssymb}
\usepackage{mathtools}
\usepackage{amsthm}
\usepackage[capitalize,noabbrev]{cleveref}
\usepackage[textsize=tiny]{todonotes}
\usepackage{mathtools}  
\usepackage[inline]{enumitem}  
\usepackage{subfig}
\setcitestyle{numbers, square}

\theoremstyle{plain}

\theoremstyle{definition}

\theoremstyle{remark}

\crefname{equation}{Eq.}{Eqs.}
\crefname{figure}{Fig.}{Figs.}
\crefname{section}{Sec.}{Secs.}
\crefname{subsection}{Sec.}{Secs.}
\crefname{theorem}{Thm.}{Thms.}
\crefname{appendix}{Appx.}{Appx.}
\crefname{lemma}{Lemma}{Lemmas}
\crefname{algocf}{Alg.}{Algs.}
\Crefname{algocf}{Algorithm}{Algorithms}
\hypersetup{
  colorlinks,
  linkcolor={blue},
  citecolor=blue,
  urlcolor=blue,
  breaklinks=true,   
  linktoc=all,
}

\newif\ifcomments
\commentstrue
\ifcomments\newcommand{\comments}[1]{#1}\else\newcommand{\comments}[1]{}\fi

\definecolor{clrgp}{rgb}{.9,0,.9}
\definecolor{red}{rgb}{.8,0,0}
\definecolor{blue}{rgb}{0,0, 0.8}
\definecolor{gray}{rgb}{0.41, 0.41, 0.41}
\definecolor{forestgreen}{rgb}{0.13, 0.55, 0.13}
\definecolor{subtle}{RGB}{152,78,163}

\newcommand{\gp}[1]{\comments{\textcolor{black}{#1}}}
\newcommand{\ekb}[1]{\comments{\textcolor{black}{#1}}}

\newcommand{\ta}[1]{\comments{\textcolor{black}{#1}}}

\newcommand{\brierscore}

\title{Deep Ensembles Work, But Are They Necessary?}

\author{%
Taiga Abe$^{*1}$
\And
E. Kelly Buchanan$^{*1}$
\And
Geoff Pleiss$^{1}$
\And
Richard Zemel$^{1}$
\And
John P. Cunningham$^{1}$ \\
$^1$Columbia University \\
$\texttt{\{ta2507,ekb2154,gmp2162,jpc2181\}@columbia.edu}$ \\
$\texttt{zemel@cs.columbia.edu}$ \\
}

\usepackage{amsmath,amsfonts,bm}

\def\eqref#1{equation~\ref{#1}}

\def\1{\bm{1}}

\def\vone{{\bm{1}}}

\def\vf{{\bm{f}}}

\def\vx{{\bm{x}}}

\DeclareMathAlphabet{\mathsfit}{\encodingdefault}{\sfdefault}{m}{sl}
\SetMathAlphabet{\mathsfit}{bold}{\encodingdefault}{\sfdefault}{bx}{n}

\newcommand{\R}{\mathbb{R}}

\DeclareMathOperator*{\E}{\mathbb{E}}
\DeclareMathOperator*{\Var}{\mathrm{Var}}
\DeclareMathOperator*{\jsd}{\mathrm{JSD}}
\DeclareMathOperator*{\entropy}{\mathrm{H}}

\usepackage{multirow}
\usepackage{enumitem}
\usepackage{wrapfig}
\begin{document}
\maketitle
\def\thefootnote{*}\footnotetext{\text{Equal contribution.}}

\begin{abstract}
Ensembling neural networks is an effective way to increase accuracy, and can often match the performance of \ta{individual} larger models. This observation poses a natural question: given the choice between a deep ensemble and a single neural network with similar accuracy, is one preferable over the other?
Recent work suggests that deep ensembles may offer \ta{distinct} benefits beyond predictive power: namely, uncertainty quantification and robustness to dataset shift.
In this work, we demonstrate limitations to these purported benefits, and show that a single (but larger) neural network can replicate these qualities.
First, we show that ensemble diversity, by any metric, does not meaningfully contribute to an ensemble's uncertainty quantification on out-of-distribution (OOD) data, but is instead highly correlated with the relative improvement of a single larger model.
Second, we show that the OOD performance afforded by ensembles is strongly determined by their in-distribution (InD) performance, and---in this sense---is not indicative of any ``effective robustness."
\ta{While deep ensembles are a practical way to achieve improvements to predictive power, uncertainty quantification, and robustness, our results show that these improvements can be replicated by a (larger) single model. }
\end{abstract}

\section{Introduction}
\label{sec:intro}
In many real-world settings, practitioners deploy ensembles of neural networks that combine the outputs of several individual models \citep[e.g.][]{szegedy2015going,kurutach2018model,yu2020mopo}.
Though training and evaluating multiple models is computationally expensive,
a wide body of research demonstrates that ensembles achieve better performance (as measured by accuracy, negative log likelihood, or a variety of other metrics) than their constituent single models, provided that these models make diverse errors  \citep{dietterich2000ensemble}.
This benefit is well-established in the literature:
theoretically proven for ensembles formed via boosting or bagging \citep{schapire1990strength,breiman1996bagging},
and demonstrated for \textit{deep ensembles} that solely rely on the randomness of SGD coupled with non-convex loss surfaces \citep{lee2015m,fort2019deep}.

Of course, ensembling is not the only way to increase performance;
one could also increase the depth or width of a single neural network.
In many settings, a single large model performs similarly to an ensemble of (smaller) models with a similar number of parameters \citep{lobacheva2020power,kondratyuk2020ensembling,wasay2020more}.
This observation poses a natural question:
are there reasons to choose a deep ensemble over a single (larger) neural network with comparable performance?

Recent research suggests that deep ensembles may be preferable to single models in 
 safety-critical applications and settings where data shifts significantly away from the training distribution.
First, \citet{lakshminarayanan2016simple} demonstrate that deep ensembles provide \emph{well-calibrated estimates of uncertainty} on classification and regression tasks.
Compared with other uncertainty quantification (UQ)  methods, ensembles offer better (i.e. less overconfident) uncertainty estimates on out-of-distribution ({OOD}) or shifted data \citep{ovadia2019can}.
Second, recent work indicates that---beyond calibration---ensemble performance (as measured by accuracy, NLL, or other metrics) also tends to be \emph{robust against dataset shift},
again often outperforming other methods in these regimes \citep{gustafsson2020evaluating}.

Intuitions in recent papers  \citep[e.g.][]{lee2015m,fort2019deep} attribute these UQ/robustness benefits to the fact that ensembles produce multiple diverse predictions, rather than a single point prediction.
If diversity does in fact explain UQ/robustness improvements, this would suggest that deep ensembles indeed offer benefits that cannot be obtained by (standard) single neural networks.
In this paper, we rigorously test hypotheses that formalize this intuition.
Surprisingly, after controlling for factors related to the performance of an ensemble's component models,
we find no evidence that having a diverse set of predictions is responsible for these purported benefits.
Put differently, we find that these UQ/robustness benefits are not unique to deep ensembles, \emph{as they can be replicated through the use of (larger) single models}.
We confirm these results for a wide variety of model architectures, as well as for \emph{heterogeneous deep ensembles} that combine multiple different neural network architectures and \emph{implicit deep ensembles} like MC Dropout \citep{gal2016dropout}, BatchEnsemble \citep{wen2020batchensemble}, and MIMO \citep{havasi2021training} (\cref{sec:app_heter_ensemble}).

\textbf{Hypothesis: ensemble diversity is responsible for improved UQ.}
Two components contribute to ensemble uncertainty estimates:
the uncertainties expressed by individual ensemble members,
and diversity among ensemble member predictions.
Recent work suggests that the diversity component is primarily responsible for better calibrated OOD uncertainty estimates, as ensemble members should agree less (i.e. offer more diverse predictions) as data shift away from the training distribution \citep{lakshminarayanan2016simple,fort2019deep,gustafsson2020evaluating}.
In contrast, we find that---after conditioning on the uncertainty of individual ensemble members---%
\emph{the level of ensemble disagreement does not statistically differ between InD and OOD data} (\cref{fig:f0}),
and thus ensemble diversity is not directly responsible for larger OOD uncertainty estimates.
\gp{Furthermore, ensemble diversity---on a per-datapoint basis---is correlated with the expected improvement we obtain by increasing model capacity (\cref{fig:perfcomp}), implying that ensemble diversity does not capture a quantity inaccessible to a single (larger) model.}

\textbf{\ta{Hypothesis: ensemble diversity is responsible for improved robustness.}}
\ta{Independent work} demonstrates a deterministic relationship between a (single) neural network's 0-1 accuracy on InD and OOD datasets \cite{taori2020measuring,miller2021accuracy},
whereby \ta{the OOD performance of a model can be predicted from its InD performance}.
It is therefore natural to ask whether having multiple diverse predictions contributes to additional OOD robustness (as suggested by \citep{fort2019deep,gustafsson2020evaluating}), beyond what is expected given performance improvements on InD data.
Our results demonstrate that deep ensembles are not ``effectively robust'' relative to single models---i.e. their OOD performance (as measured by accuracy, NLL, Brier score, and calibration error) follows the same deterministic relationship to InD performance as single models (\cref{fig:ltrend_metrics}).
Therefore, ensemble diversity does not yield additional robustness over what standard single networks achieve.

\textbf{Implications.}
Overall, this paper does not disagree with prior claims about the benefits of deep ensembles relative to an ensemble's component models.
Indeed, in our experiments we confirm that ensembling is a convenient mechanism to improve predictive performance, UQ, and robustness relative to this baseline.
At the same time, our results also indicate that---after controlling for individual model uncertainty and InD performance---
ensembles do not obtain UQ/robustness benefits beyond what can already be obtained from the properties of an appropriately chosen single model.

\section{Related work}
\label{sec:relatedwork}
Ensembling is an established technique to improve generalization \citep[e.g.][]{schapire1990strength,perrone1992networks,Domingos1997WhyDB,opitz1999popular}, where the predictions of multiple models are aggregated to reach a consensus.
It is well established that diversity amongst ensemble members is necessary to improve performance \citep{dietterich2000ensemble}.
This diversity can be achieved through many means.
Randomization approaches introduce diversity by training each model on a random subset of data \citep{breiman1996bagging} or a random subset of features \citep{breiman2001random}.
Alternatively, boosting approaches \citep{freund1995boosting,friedman2001greedy} achieve diversity by manipulating the weighting of training data.
Other methods include using a diverse set of model classes \citep[e.g.][]{caruana2004ensemble} or joint training objectives \citep[e.g.][]{munro1997competition}.

\textbf{Ensembles of neural networks.} %
Historically, neural network ensembles have relied on a variety of mechanisms to introduce diversity \citep[e.g.][]{hansen1990neural,perrone1992networks,moghimi2016boosted,zaidi2020neural}.
Recently, diversity is often obtained by training multiple copies of the same neural network architecture with different intializations and minibatch orderings,
as the inherent randomness of SGD has been shown to introduce a sufficient amount of diversity in these (non-convex) models \cite{lee2015m,goodfellow2016deep,fort2019deep}.
Importantly, this approach can exploit parallel computation \cite{lakshminarayanan2016simple},
because none of the ensemble members depend on one another.

\textbf{Deep ensembles for predictive uncertainty.}
It has been suggested that ensembles of neural networks not only improve accuracy but also estimates of predictive uncertainty \cite{lakshminarayanan2016simple}.
Some research aims to connect ensembles and Bayesian neural networks,
suggesting that these improved uncertainty estimates are the result of performing approximate Bayesian model averaging \cite{gal2016dropout,wilson2020bayesian}.
Although prior work has described shortcomings in the uncertainty estimates derived from deep ensembles \cite[e.g.][]{liu2020simple,ciosek2019conservative,he2020bayesian,osband2021epistemic}, 
they remain a gold standard in high risk and safety critical settings %
\cite[e.g.][]{ovadia2019can,gustafsson2020evaluating,tran2022plex}.  

\textbf{Deep ensembles and robustness.}
Robustness is the ability to maintain good accuracy and calibration under conditions of distributional shift.
Deep ensembles outperform other approaches in maintaining both accuracy and calibration on OOD data \cite{ovadia2019can,gustafsson2020evaluating}, although their limitations have also been demonstrated \cite{kumar2021calibrated,rahaman2020uncertainty}.
This robustness is attributed to the diversity between ensemble members \citep{fort2019deep}.

\textbf{Other related work.}
Recent work investigates whether it is possible to achieve the benefits of an ensemble with reduced computation during training and/or test time \cite{huang2017snapshot,maddox2019simple,wen2020batchensemble,havasi2021training}.
Additionally many works have proposed numerous diversity metrics for ensembles similar to those we examine here \citep[e.g.][]{krogh1995cross,masegosa2020learning,andres2022diversity}.

\section{Setup}
\label{sec:setup}

Consider multiclass classification: inputs $\vx \in \R^D$ with targets $y \in [1,\dots,C]$, where $D$ is the number of features and $C$ is the number of classes.
We assume that we have access to $M$ distinct neural networks $\vf_1, \ldots, \vf_M$,
where each model $\vf_i : \R^D \to \Delta^C$ maps an input to the $C$-class probability simplex.
We will primarily focus on the common case of {\bf homogeneous ensembles}, where $\vf_1, \ldots, \vf_M$ represent the same neural network architecture and training procedure, relying on the inherent randomness of initialization and SGD to produce diverse models (see Sec.~\ref{sec:relatedwork} for a broad discussion).
However, in \cref{sec:heterogeneous} we will also consider {\bf heterogeneous ensembles} where $\vf_1, \ldots, \vf_M$ represent different architectures or training procedures, \ekb{and {\bf implicit ensembles}, where $\vf_1, \ldots, \vf_M$ are approximated by changes to a single model \cite{gal2016dropout,havasi2021training,wen2020batchensemble}}.
Throughout the paper, we will also represent these member networks as a discrete distribution of models:
$p(\vf) = \text{Unif.} [ \vf_1, \ldots, \vf_M ] $.
The ensemble prediction $\bar \vf(\vx)$ is given by the arithmetic mean of the ensemble member \emph{probabilities}:\footnote{
    While it is also possible to average the logits (log probabilities) of each model,
    we note that probability averaging is far more common in the literature
    \citep[e.g.][]{lakshminarayanan2016simple}.
}
\begin{equation}
    \textstyle{
        \bar \vf(\vx) = \E_{p(\vf)}[\vf(\vx)] = \frac{1}{M} \sum_{i=1}^M \vf_i(\vx)
    }
    \label{eqn:ensemble}
\end{equation}

\textbf{Metrics for ensemble diversity.}
Two metrics of ensemble diversity are
\begin{enumerate*}[label=(\arabic*)]
    \item variance \citep[e.g.][]{kendall2017uncertainties}, and
    \item Jensen-Shannon divergence \citep[e.g.][]{lakshminarayanan2016simple,fort2019deep}.
\end{enumerate*}
Mathematically, they are (respectively) defined as:
\begin{align}
    \Var_{p(\vf)} \left[ \vf(\vx) \right] &= {\textstyle \sum_{i=1}^C } \Var_{p(\vf)} \left[ f^{(i)} (\vx) \right], \quad
    \label{eqn:variance}
    \jsd_{p(\vf)} \left[ y \! \mid \! \vf(\vx) \right] = \entropy \bigl[ y \! \mid \! \bar \vf(\vx) \bigr] -
        \E_{p(\vf)} \bigl[ \entropy \left[ y \! \mid \! \vf(\vx) \right] \bigr]
\end{align}
\ta{where $f^{(i)}$ refers to the probability assigned by a model to the $i$-th output class, and H is the entropy. Both metrics} are always positive and minimized when the predictions from ensemble members are the same, i.e. not diverse.

\textbf{Models and training datasets.}
We reuse and train a variety of neural networks on two benchmark image classification datasets:
{\bf CIFAR10} \citep{krizhevsky2009learning} and {\bf ImageNet} \citep{deng2009imagenet}.
In particular, we include the 137 CIFAR10 models trained by \citet{miller2021accuracy}, corresponding to 32 different architectures each trained for 2-5 seeds; as well as the ``standard" 78 ImageNet models curated by \citet{taori2020measuring}, each corresponding to a different architecture trained for 1 seed. To form homogeneous ensembles, we additionally train 10 network architectures on CIFAR10
and three on ImageNet.
We train 5 independent instances of each model architecture, where each instance differs only in terms of initialization and minibatch ordering.
We form homogeneous deep ensembles by combining 4 out of the 5 random seeds.
From this process, we can consider 5 single model replicas and 5 ensemble replicas for each model architecture.
Unless otherwise stated, ensembles are formed following \cref{eqn:ensemble}.

\textbf{OOD datasets.}
A majority of our analysis compares deep ensembles on InD versus OOD test data.
To that end, we consider three different catagories of OOD datasets as suggested by \citep{miller2021accuracy}:
\emph{Shifted reproduction datasets.}
        This category includes the {\bf CIFAR10.1} and {\bf ImageNetV2} datasets \citep{recht2019imagenet},
        both of which were collected and labeled following the same curation processes of the original CIFAR10 and ImageNet datasets, respectively.
        Neural networks (trained on the original datasets) tend to achieve worse performance on these new test sets.
\emph{Alternative benchmark datasets.}
        The {\bf CINIC10} dataset \citep{darlow2018cinic} shares the same classes as CIFAR10 but uses images drawn and downsampled from the ImageNet dataset.
        Because ImageNet and CIFAR10 images were collected using different curation procedures, models trained on CIFAR10 tend to achieve worse performance on CINIC10.
\emph{Synthetically corrupted datasets.}
        The {\bf CIFAR10C} and {\bf ImageNetC} datasets \cite{hendrycks2018benchmarking},
        apply synthetic perturbations to CIFAR10 and ImageNet images (e.g. Gaussian blur, fog effects, etc.).
        Due to their synthetic nature, these datasets offer shifts of various intensity (e.g. mild blur versus heavy blur).
        We relegate most of our analysis of these datasets to the Appendix.

\begin{figure*}[!t]
\centering
\includegraphics[width=1.0\textwidth]{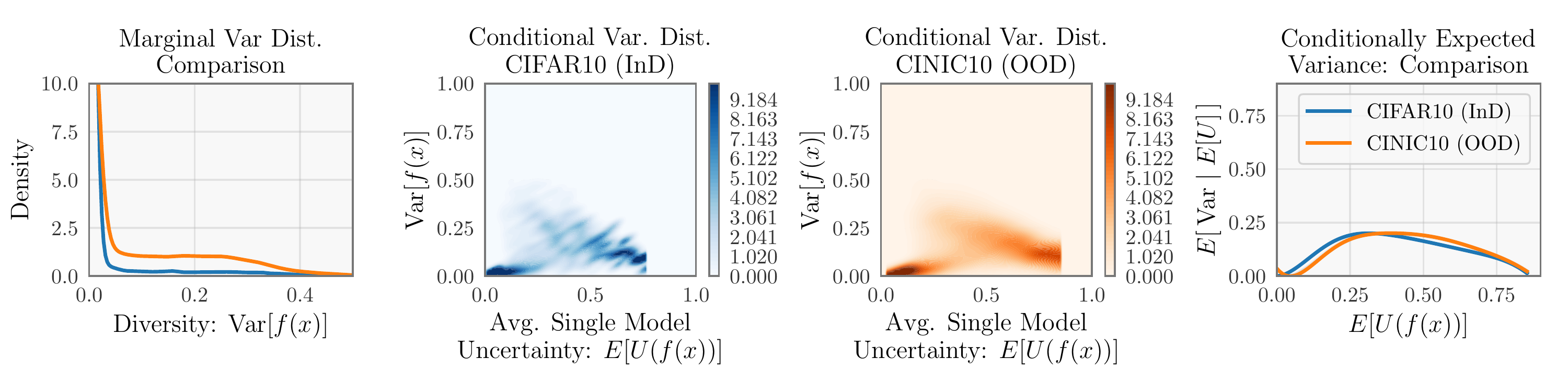}
\begin{tikzpicture}[      
        every node/.style={anchor=south west,inner sep=0pt},
        x=1mm, y=1mm,
      ]   
     \node (fig1) at (0,0)
       {\includegraphics[width=1\textwidth]{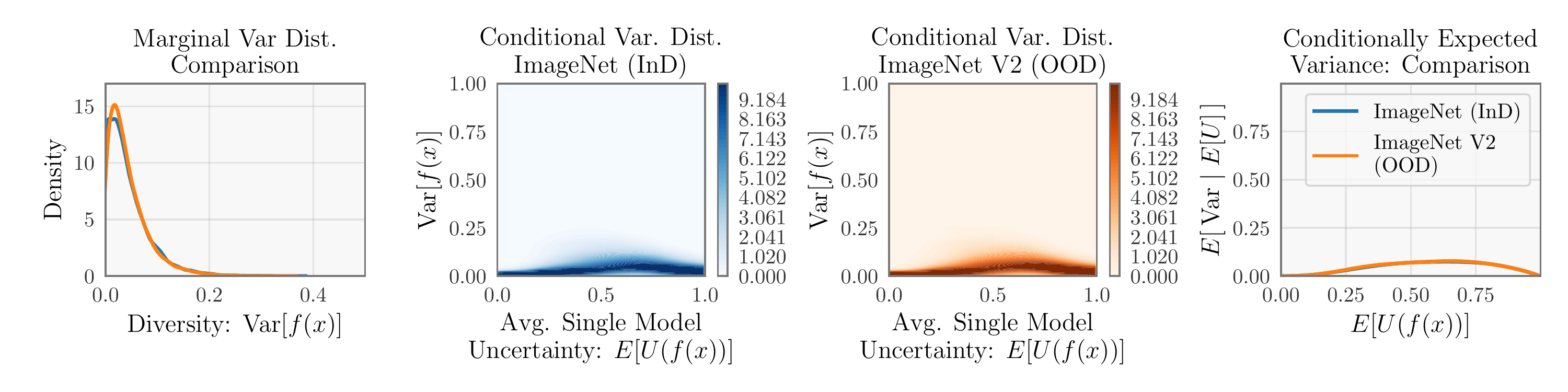}};
     \node (fig2) at (19,17)
       {\includegraphics[width=0.09\textwidth]{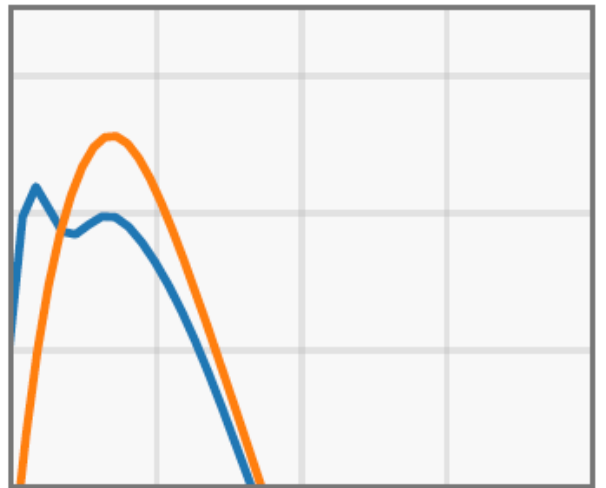}};  
\end{tikzpicture}

\caption{\textbf{Ensemble diversity does not yield better OOD uncertainty quantification, after controlling for average single model uncertainty.} 
Panels compare ensemble variance \ta{($\Var[ \vf(\vx)  ]$)} on InD (blue) vs. OOD (orange) data.  \ekb{The top row represents the variance for ensembles composed of 5 WideResNet 28-10 \cite{zagoruyko2016wide} networks evaluated on CIFAR10 and CINIC10, and the bottom row represents the variance for ensembles of 5 AlexNets, evaluated on ImageNet and ImageNetV2. 
The left column shows that, consistent with previous results, deep ensembles express higher variance predictions on OOD vs. InD data. 
The middle columns show} \ta{ $p(\Var\mid \mathbb{E}[U])$ (arguments suppressed for clarity)
}\ekb{
for InD (second column) and OOD data (third column). Suprisingly, we find that these conditional distributions are extremely similar.
In the right columns,}
\ta{we further show the}
\ekb{similarity of these conditional distributions (InD and OOD) using the conditional expectation $\E[\Var \mid \E[U]]$, estimated with kernel ridge regression. For experimental details, see \cref{sec:appconddiv}. }}
\label{fig:f0}
\end{figure*}

\section{Hypothesis: ensemble diversity is responsible for improved UQ}
\label{sec:uncertainty}

The ability of deep ensembles to produce higher estimates of uncertainty on OOD data has been attributed to ensemble diversity \citep{lakshminarayanan2016simple,fort2019deep,wilson2020bayesian}.
 In particular, ensemble diversity is hypothesized to increase on OOD data, where
one would expect that OOD predictions from individual ensemble members are less constrained by their shared training data
\citep{lakshminarayanan2016simple}. 
This hypothesis is attractive because it suggests that deep ensembles offer an additional mechanism for uncertainty quantification beyond what is afforded by any single model.
In this section, we test this hypothesis by quantifying the contribution of ensemble diversity to a deep ensemble's total predictive uncertainty on both InD and OOD data.

\subsection{Metrics for ensemble diversity\label{sec:metricepiun}}
\ta{Common metrics for ensemble diversity provide interpretable decompositions of uncertainty: \emph{ensemble uncertainty $=$ ensemble diversity $+$ average single model uncertainty}.}
For example, 
if we use variance (Eq.~\ref{eqn:variance}) as a metric for ensemble diversity \citep{kendall2017uncertainties,gustafsson2020evaluating}, then we show ensemble uncertainty can be decomposed as:
\begin{align}
  \overbracket{
    \: U \left( \bar \vf(\vx) \right) \:
  }^{\text{ens. uncert.}}
  = \overbracket{
    \: \Var_{p(\vf)} \left[ \vf(\vx) \right] \:
  }^{\text{ens. diversity}}
  + \overbracket{
    \E_{p(\vf)} \left[  U\left( \vf(\vx) \right) \right] \:
  }^{\text{avg. single model uncert.}}.
  \label{eqn:variance_uncertainty}
\end{align}
where $\vf(\vx) \in \Delta^C$ is a probabilistic prediction,  and $U( \vf(\vx) )$ is a quadratic notion of uncertainty:
\[
    U \left( \vf(\vx) \right) \triangleq 1 - \textstyle{\sum_{i=1}^C }\bigl[ p(y = i \mid \vf(\vx)) \bigr]^2.
\]
See derivation in \cref{sec:uncertdecomp}. Intuitively, $U$ will be small when most probability is placed on a single class,
and will be large when probability is distributed amongst classes. See \cref{sec:uncertdecomp} for analogous results with Jensen Shannon divergence as the diversity metric (Eq.~\ref{eqn:variance}). 
Based on our hypothesis,
ensemble diversity ($\Var$ in Eq.~\ref{eqn:variance_uncertainty}) should increase on OOD data \textit{independently} of average single model uncertainty ($\E[ U ]$). 
In other words, given \textit{any} level of $\E[ U ]$, we would expect more ensemble diversity for OOD data than InD data.

\subsection{Experiment: InD vs OOD ensemble diversity\label{sec:diversity_exp}}

We test 10 different ensembles of size $M=5$ trained on CIFAR10, and three ensembles trained on ImageNet.  
We evaluate these ensembles on their respective InD (CIFAR10, Imagenet) and OOD (CIFAR10.1, CINIC10, CIFAR10C, ImageNet V2, ImageNetC) test sets. 
\ekb{In \cref{fig:f0}, we analyze the variance of two of these deep ensembles, evaluated on CIFAR10 vs CINIC10 (top row) and ImageNet vs ImageNetV2 (bottom row), see \cref{sec:appconddiv} for a complete set of results. The left panel of \cref{fig:f0} } \ta{shows the distribution $p(\mathrm{Var})$ for InD and OOD data.} \ekb{Ensembles tend to express higher variance on OOD data than InD data; a finding 
consistent with previous work 
\citep{lakshminarayanan2016simple,fort2019deep}.}
However, we emphasize this result is not sufficient to directly attribute UQ improvements to ensemble diversity. 

\textbf{Controlling for single model uncertainty. }
A different picture emerges when we control for single model uncertainty.
\cref{fig:f0} (middle) shows histograms of $p(\Var\mid \mathbb{E}[U])$ i.e. the ensemble variance \textit{conditioned on} average single model uncertainty as given by \cref{eqn:variance_uncertainty}.
Surprisingly, we see that the OOD and InD conditional distributions are very similar.
We \ta{further study} this similarity in \cref{fig:f0} (right), which plots expected ensemble variance conditioned on average single model uncertainty: \ta{$\E[\,{ \Var \mid \mathbb{E}[U]}\,]$}.
Far from what our hypothesis would suggest (i.e. higher OOD diversity across all levels of average single model uncertainty)
we observe that the conditional \ta{expectation} of ensemble diversity on InD vs OOD data is nearly identical. 
In \cref{sec:appconddiv} (\cref{fig:percentincrease_Var_cifar}-\cref{fig:percentincrease_imagenet}), we \ta{offer statistical validation of these observations, and further} demonstrate that this phenomenon holds across various architectures, InD, and OOD datasets. \ta{In all cases}, the difference between the InD and OOD expected variance is only a few percentage points, and/or not statistically significant. 

\textbf{Understanding the relationship between ensemble diversity and average single model uncertainty.} By controlling for average single model uncertainty, we see that ensemble diversity does not differ significantly for InD versus OOD data. 
In turn, 
these results imply that the 
InD/OOD difference
we see in \cref{fig:f0} (left) must be due entirely to a change in the distribution of \textit{average single model uncertainty}, \ta{$p(\mathbb{E}[U])$}. 
From these results, we can conclude that surprisingly, the UQ benefits of ensemble diversity are dictated by 
the corresponding average single model uncertainty.
In \cref{appx:avg_plots} we plot the differences in $p(\mathbb{E}[U])$ that drive the changes in ensemble diversity observed in \cref{fig:f0} (left).

\begin{figure*}[!t]
\centering
\subfloat{\includegraphics[width=0.5\linewidth]{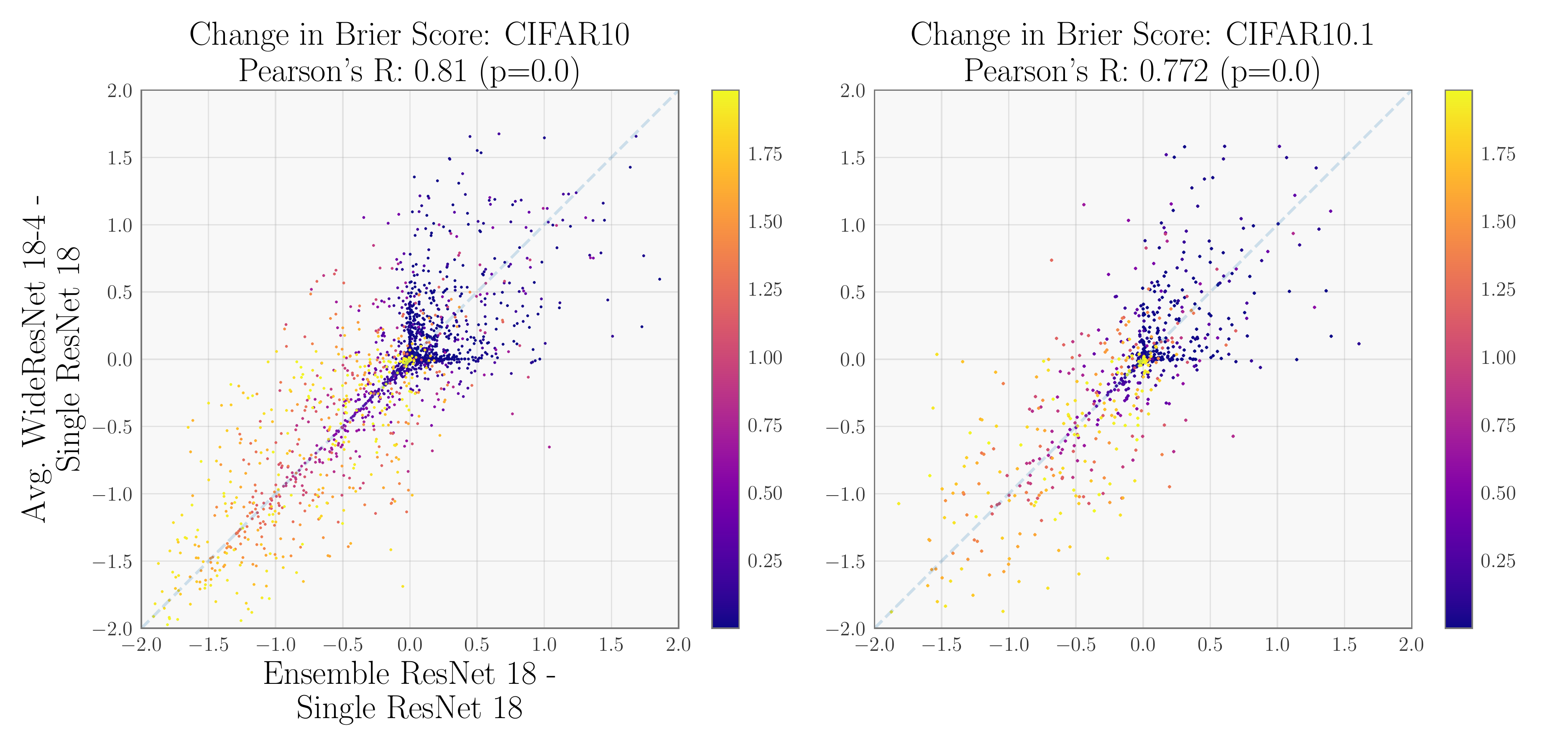}}
\subfloat{\includegraphics[width=0.5\linewidth]{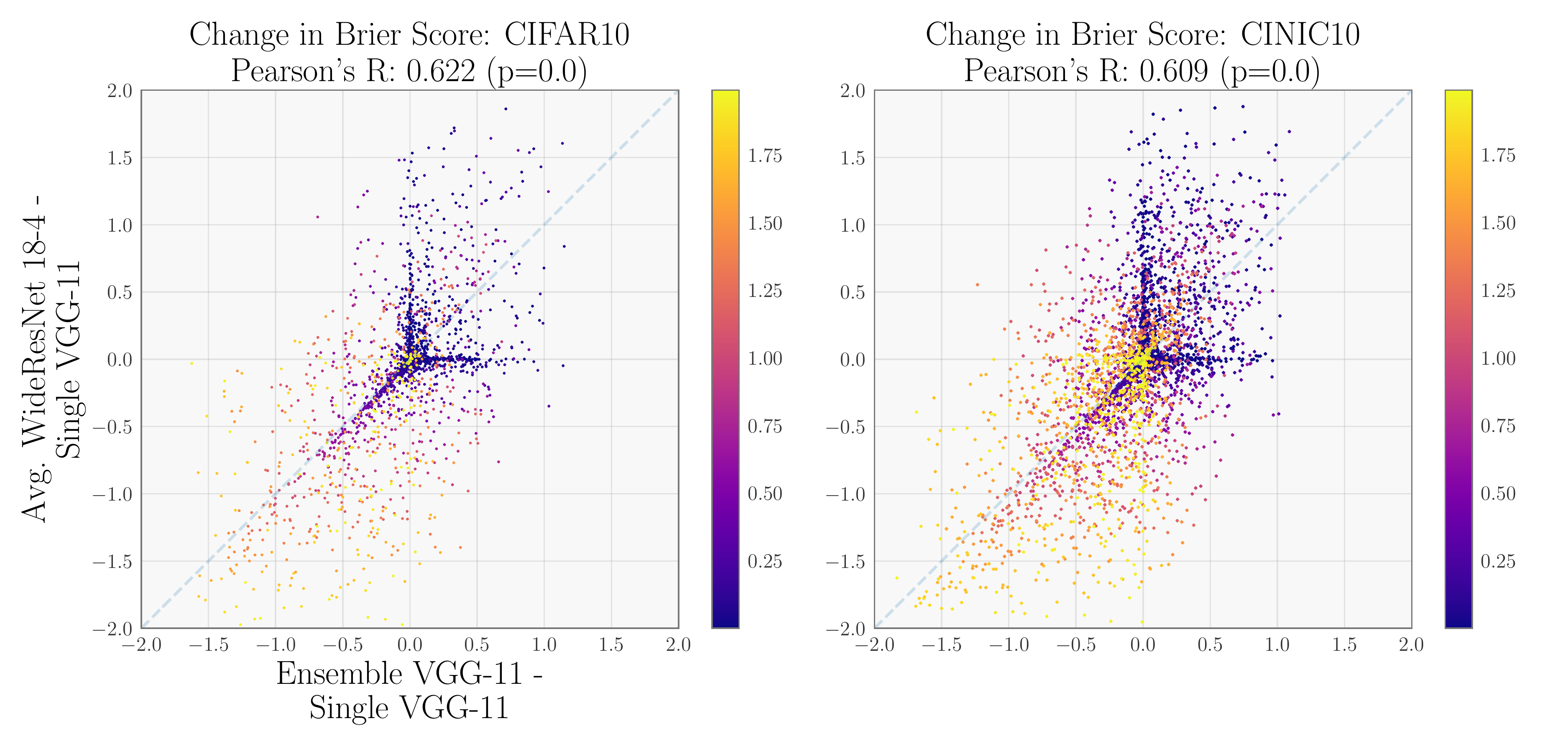}}
\caption{\textbf{Ensemble diversity is \ta{meaningfully} correlated with the expected improvements from increasing model capacity}. 
(Left) Panels illustrate the per-datapoint gains in Brier score over a single ResNet 18 model by either forming a deep ensemble of ResNet 18 models (x-axis), or by increasing single model capacity, here with a WideResNet 18-4 (y-axis).
The ResNet-18 ensemble and WideResNet 18-4 achieve nearly identical  performance and strongly correlated improvements on both CIFAR10 and CIFAR10.1.
Colors indicate the Brier score achieved by the single ResNet 18 model on each datapoint. 
(Right) We repeat the experiment 
for CIFAR10/CINIC10, showing the gains in Brier score over a VGG11 model, using either an ensemble of VGG11, or a WideResNet 18-4 model.
Improvements are indistinguishable from relevant controls, and corresponding model accuracies are well matched, as shown in \cref{sec:appmmd}.
}
\label{fig:perfcomp}
\end{figure*}

\subsection{What does ensemble diversity actually measure?}
\label{sec:correlated_improvemenets_pp}

Our analysis above shows that ensemble diversity is not directly responsible for the improved OOD uncertainty estimates offered by ensembles.
To begin to understand why this might be the case,
it is useful to consider the link between ensemble diversity and performance.
It has long been established that diversity amongst ensemble members is a necessary and sufficient condition for the superior performance of ensembles \citep[e.g.][]{dietterich2000ensemble}.
To demonstrate this, consider any strictly convex loss function, such as negative log likelihood (NLL) or the multiclass Brier score (B) \citep{brier1950verification}:
\begin{align}
    \text{NLL}(\vf(\vx), y) \triangleq - \log \left( f^{(y)} (\vx)\right), \quad
    \mathrm{B}(\vf(\vx), y) \triangleq \Vert \vf(\vx) - \vone_y \Vert_2^2.
    \label{eqn:brier}
\end{align}
(Here, $\vone_y$ represents a one-hot encoding of $y$.)
Recall that the ensemble prediction $\bar \vf(\vx)$ is the average of all model predictions (i.e. $\E_{p(\vf)}[\vf(\vx)]$).
By Jensen's inequality:
\begin{equation}
    \begin{split}
        \text{NLL}(\bar \vf(\vx), y)  \leq \E_{p(\vf)} \left[ \text{NLL}( \vf(\vx), y  ) \right], \quad
        \mathrm{B}(\bar \vf(\vx), y) \leq \E_{p(\vf)} \left[ \mathrm{B}( \vf(\vx), y  ) \right]
    \end{split}
    \label{eqn:jensen}
\end{equation}
In other words, the performance of the ensemble (as measured by NLL or Brier score) must be better than the average performance of ensemble members.
Because both NLL and Brier score are strictly convex, the Jensen gap in \cref{eqn:jensen} will grow as $p(\vf)$ becomes less constant, or more ``diverse.''
In particular, the Jensen gap for Brier score is exactly equal to the ensemble variance (Eq.~\ref{eqn:variance}):
\begin{equation}
    \mathrm B(\bar{\vf}(\vx),y) - \E_{p(\vf)}[ \mathrm B(\vf(\vx),y)] = \Var_{p(\vf)}[\vf(\vx)].
    \label{eqn:brier_variance}
\end{equation}
(Similar results are well known in the regression context---\citep[e.g.][]{krogh1995cross,masegosa2020learning}---see \cref{sec:app_brier_score_uncertainty_decomposition} for a short derivation).
In other words, $\Var_{p(\vf)}[\vf(\vx)]$ measures the expected predictive improvement we obtain through ensembling.
We can use these results to investigate our UQ findings.
Hypothetically, if $\Var_{p(\vf)}[\vf(\vx)]$ were also responsible for improving UQ,
this would imply that the performance gains from ensembling are somehow fundamentally different than the performance gains from increasing a single model's capacity, as the latter can hurt uncertainty estimates \citep{guo2017calibration}.
However, in the next section we demonstrate that these two methods of increasing performance are in fact correlated.

\subsection{Ensembling versus increasing model capacity}
In \cref{fig:perfcomp}, we compare the expected per-datapoint performance improvement gained through ensembling (x-axis) to the performance improvement gained through increasing model capacity (y-axis).
Specifically, we compare an ensemble of 4 CIFAR10 models (ResNet18) with a single large model (WideResNet-18-4).
The ensemble and the large single model achieve comparable Brier Score: $0.084 \pm 0.002$  on the InD test dataset and $0.210\pm0.002$ on the CIFAR10.1 OOD dataset.
In \cref{fig:perfcomp} (left), we plot the Brier score of the ensemble versus the large model on a per-datapoint level, \ta{depicting the \textit{improvement correlation} across the dataset}.

Surprisingly, we find that increasing model capacity and ensembling yield very similar performance improvements \emph{on most datapoints}.
The ensemble improvements and large model improvements have a Pearson's correlation of 0.81 on the InD test set.
Importantly, we see that this correlation is preserved even on OOD data (Pearson's correlation: 0.76).
We replicate this result for a different ensemble/larger model pair (VGG-11 ensemble versus WideResNet-18-4) that again have nearly identical InD and OOD performance: $0.093\pm0.004$ CINIC10 InD Brier Score; $0.48\pm0.02$ CINIC10 OOD Brier Score (\cref{fig:perfcomp}, right).
\ta{We compare each improvement correlation in \cref{fig:perfcomp} to relevant controls, and ensure comparable accuracies (\cref{sec:appmmd}). In all cases we find that improvements are as similar as we might expect if comparing two performance matched ensembles, or two single models.}
This result is unexpected, because the ensemble and the large model represent two distinct architectures (ResNet versus WideResNet) and two different modes of training (independent training of separate models versus training one large model).
Recalling the relationship between ensemble diversity and relative performance gains,
these results suggest that \emph{ensemble diversity \ta{estimates} the improvement we should expect by increasing model capacity.}
We conclude that, with regards to UQ and performance improvements, ensemble diversity offers no significant benefit over what can be obtained with single models.

\subsection{Implications for uncertainty estimation}
\label{sec:implications_epis_alea}
\textbf{Epistemic vs. aleatoric uncertainty.} Uncertainty is often categorized as coming from one of two components \cite[e.g.][]{hullermeier2021aleatoric}. The \emph{epistemic} component is said to capture uncertainty due to a limited number of observations, or uncertainty that the model accurately and uniquely captures the ground truth labeling process.
Apparently, it can be reduced by collecting more data. 
In contrast, the \emph{aleatoric} component is described as capturing the inherent ambiguity in the data (e.g. a blurry image) and is considered to be irreducible noise.
In decision making applications such as active learning \cite{settles2009active,gal2017deep} or model-based reinforcement learning \citep{kurutach2018model,yu2020mopo}, this uncertainty decomposition is employed to identify informative datapoints for our model to sample next \citep{depeweg2018decomposition}. 
Previous work has interpreted ensemble diversity as in \cref{eqn:variance} as epistemic uncertainty \citep{malinin2018predictive,gustafsson2020evaluating,yu2020mopo}, with average single model uncertainty in \cref{eqn:jsd_uncertainty,eqn:variance_uncertainty} identified as aleatoric uncertainty correspondingly \cite{smith2018understanding}.
Our results in \cref{fig:f0} demonstrate that there is a limitation to this interpretation,
as we would expect more ensemble variance (the proxy for epistemic uncertainty) for OOD data than for InD data, independent of single model uncertainty (the proxy for aleatoric uncertainty).
We therefore suggest caution when using ensembles to differentiate sources of uncertainty in downstream applications.

\textbf{Bayesian perspective.}  Bayesian model averaging, or BMA 
integrates predictions against a posterior distribution over models.
Given training data $\mathcal{D}$, BMA forms the prediction
$
p(y\mid \vx, \mathcal{D}) = \int \vf(\vx) \: p(\vf \mid \mathcal{D}) \: \mathrm d \! \vf.
$
The advantage of BMA is the ability to consider all possible predictions given a prior and conditioned on training data, thereby mitigating the risk in estimating the ``true'' model from limited data.
A recent line of work argues that modern deep ensembles (unlike classic ensembles---see \citet{minka2000bayesian}) can be viewed as approximate BMA \citep{hoffmann2021deep,wilson2020bayesian}, although we also note that concurrent work emphasizes differences between deep ensembles and Bayesian inference in the infinite width limit \citep{NEURIPS2020_0b1ec366}. 
Our results in \cref{fig:f0} identify a limitation of ensembles as approximate Bayesian inference.
The posterior predictive distribution should express higher variance for OOD data than InD data, which is not the case for the deep ensemble predictive distribution.
In \cref{sec:gp_uncertainty}, we demonstrate that exact Bayesian inference does yield higher OOD posterior variance, even after conditioning on observational noise.
We emphasize that our results neither agree nor disagree with the BMA interpretation of ensembling.
Rather they suggest that ensemble members should not be interpreted as true posterior samples,
and that (as with many approximate Bayesian methods) the ensemble approximation to BMA is biased.

\section{Hypothesis: ensemble diversity is responsible for improved robustness} %
\label{sec:ltrend}

Beyond uncertainty quantification, ensembles have been shown to often achieve better predictive performance than single networks (as measured by 0-1 accuracy, NLL, or Brier score) on OOD or shifted datasets \citep{lakshminarayanan2016simple,ovadia2019can,gustafsson2020evaluating}.
In this section, we test the hypothesis that ensemble diversity improves robustness over what single neural networks can offer.

\begin{wrapfigure}{R}{0.40\textwidth}
\centering
 \vspace{-25pt}
\includegraphics[width=0.9\linewidth]{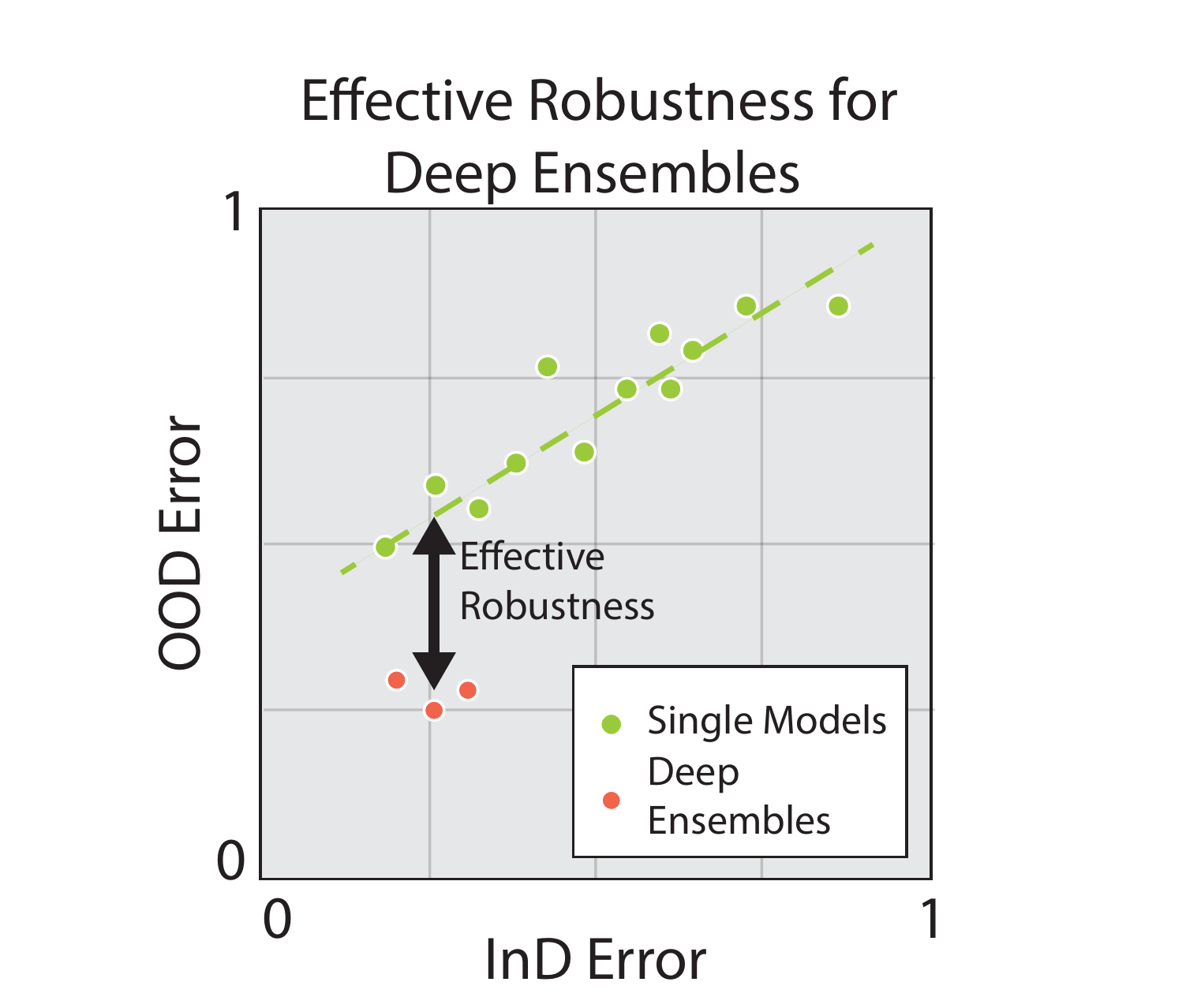}%

\caption{\textbf{Cartoon of effectively robust deep ensembles (what we want).}
}
\label{fig:ltrend_schematic}
\vspace{-10pt}
\end{wrapfigure}
\subsection{Effective robustness}

We use the concept of ``effective robustness'' as introduced by \citet{taori2020measuring}.
These authors note that there is often a deterministic relationship between a neural network's accuracy on InD data and its accuracy on an OOD dataset (green line in \cref{fig:ltrend_schematic}).
In other words, any improvements in OOD performance can be entirely explained by improvements in OOD performance.
A model is considered to be \emph{effectively robust} only if it achieves better OOD accuracy than what is predicted by its InD accuracy.
In general, there are very few neural networks or training procedures that exhibit effective robustness against any OOD dataset \citep{taori2020measuring,miller2021accuracy}.
To measure the role that ensemble diversity plays in robustness, we quantify to what extent deep ensemble OOD performance can be explained by InD performance (as measured by the deterministic relationship derived from single models). 
If the performance of deep ensembles follows the same deterministic relationship, then deep ensembles are not effectively robust (i.e. multiple diverse predictors offer no additional robustness over what a single neural network provides).

\subsection{Experiment: measuring effective robustness of deep ensembles across metrics}

\begin{figure*}[!tb]
\centering
\subfloat{\includegraphics[width=1\textwidth]{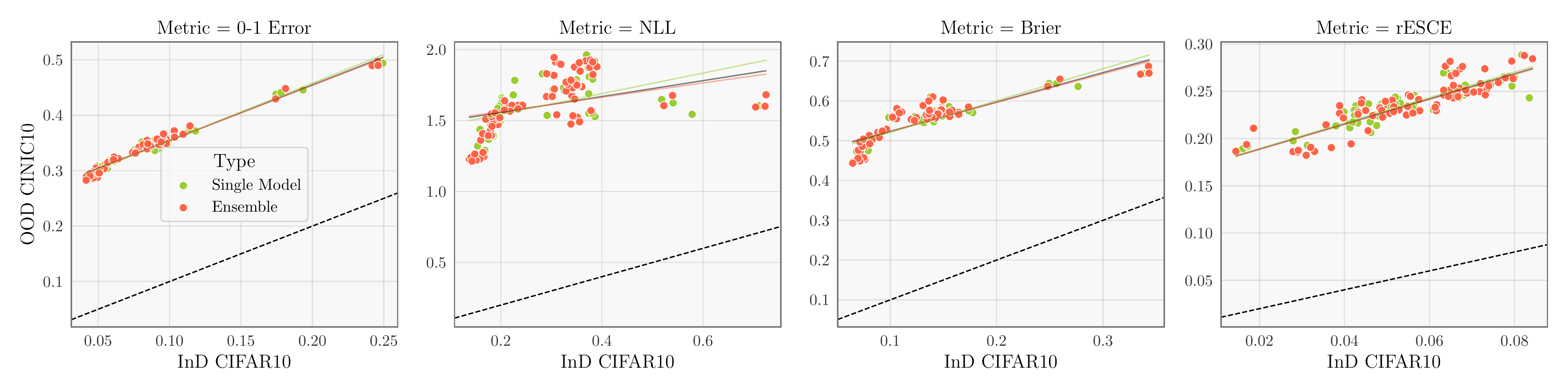}\label{fig:ltrend_cinic10}}
\hfill
\vspace{-0.2in}
\subfloat{\includegraphics[width=1\textwidth]{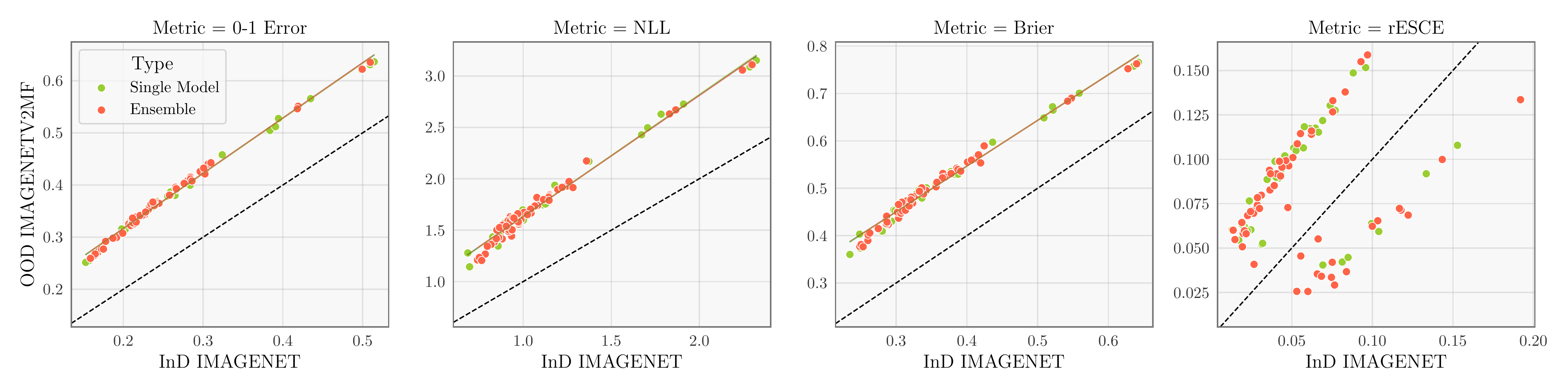}\label{fig:ltrend_imagenetv2}}
\caption{\textbf{Deep ensembles are not ``effectively robust" across a variety of performance metrics}. 
Panels illustrate InD vs OOD performance metrics, from left to right: 0-1 Error, NLL, Brier Score, and rESCE. The model types considered are single models and ensembles. 
Linear trend lines are shown in solid lines, and black dotted lines indicate perfect robustness.  We find that, conditioned on InD performance, ensembles offer no better OOD performance than single models.
See \cref{sec:additional_ltrend} for additional corruptions.}
\label{fig:ltrend_metrics}
\end{figure*}

\textbf{Ensembles are not effectively robust with respect to 0-1 accuracy.}
Following \citet{miller2021accuracy}, we measure the InD and OOD error for all the models described in \cref{sec:setup}.
The top left of \cref{fig:ltrend_metrics} compares the error of models on CIFAR10 (InD) versus CINIC10 (OOD),
and the bottom left plot compares the error of models on ImageNet (InD) versus ImageNetV2 (OOD).
From these plots, we observe several trends.
In agreement with \citet{taori2020measuring} and \citet{miller2021accuracy}, we observe that single models (green dots) follow a colinear relationship for InD versus OOD accuracy.
Additionally, we
find that \emph{ensembles (orange dots) do not deviate from this colinear InD/OOD relationship.}
In \cref{sec:ltrend_r2_tables}, we evaluate the quality of these linear trends.
In particular, we fit separate linear trend lines for individual models and deep ensembles.
All trend lines achieve correlations of $R>0.84$, and their coefficients only differ by $1\%$ at most.
This suggests that, after controlling for InD accuracy, the OOD accuracy of ensembles is nearly identical to that expected of single models.
(See \cref{sec:additional_ltrend} for CIFAR10.1/CIFAR10C/ImageNetC results.)

\textbf{Ensembles are not effectively robust with respect to NLL or Brier score.}
Although deep ensembles are not effectively robust in terms of predictive accuracy, many of their robustness benefits have been reported in terms of probabilistic metrics, such as NLL or Brier score \citep{ovadia2019can}.
We therefore extend our investigation of deep ensemble effective robustness to these metrics. 
\cref{fig:ltrend_metrics} (middle left) plots the InD NLL and OOD NLL of various ensembles and single models.
To the best of our knowledge, this is the first time that the effective robustness experiments of \citet{taori2020measuring} and \citet{miller2021accuracy} have been extended to metrics other than 0-1 accuracy.
We observe that the relationship between InD NLL and OOD NLL is not as linear as the accuracy trend.
Nevertheless, we observe no discernible difference between the performance of single networks and ensembles  (see \cref{sec:ltrend_r2_tables} for a quantitative analysis).
We observe a similiar phenomenon when we plot InD versus OOD Brier score (\cref{fig:ltrend_metrics}, middle right)---%
ensembles and single models obtain similar OOD Brier score, after controlling for InD Brier score.
Our key conclusion is that deep ensembles fail to demonstrate effective robustness when evaluated on probabilistic performance metrics, just as they do with 0-1 accuracy.
(See \cref{sec:additional_ltrend} for CIFAR10.1/CIFAR10C/ImageNetC results.)

\textbf{Ensembles do not offer effectively robust calibration.}
We also compare InD and OOD calibration for various single models and ensembles.
We consider various metrics for measuring and comparing calibration used throughout the literature.
Expected Calibration Error (ECE) \citep{naeini2015obtaining} is a standard metric for measuring calibration of neural networks.
As we show in \cref{sec:app_ece_all}, there is little correlation between a single model’s InD ECE and OOD ECE, which precludes any discussion of “effective robustness” using this metric.
Conversely, \cref{fig:ltrend_metrics} depicts a strong correlation between a model’s InD/OOD square root of the Expected \emph{Squared} Calibration Error (rESCE) \citep{degroot1983comparison,murphy1977reliability}, which appears in a common decomposition of the Brier score  \citep{brocker2009reliability}. 
We therefore expect that any InD/OOD trend for the rESCE should be qualitatively similar to the InD/OOD trends observed for Brier score.
In \cref{fig:ltrend_metrics} (top right), we observe a linear trend relating the CIFAR10 (InD) and CINIC10 (OOD) rESCE of single models.
The rESCE of the ImageNet models, \cref{fig:ltrend_metrics} (bottom right), follows a bimodal trend, where---depending on the model architecture---InD rESCE is correlated with either low or high OOD calibration.
Nevertheless, for both datasets we find that ensembles do not achieve better OOD calibration that single models with similar InD calibration.
(See \cref{sec:additional_ltrend} for CIFAR10.1/CIFAR10C/ImageNetC results.)

\subsection{Heterogeneous and implicit ensembles}
\label{sec:heterogeneous}
From the previous results, it is clear that---by many metrics---ensembling multiple copies of the same model architecture confers no additional robustness over single models.
A natural question is whether we could achieve more robustness by ensembling different model architectures together.
To test this hypothesis, we repeat the same robustness experiments with \emph{heterogeneous ensembles}: ensembles that combine multiple architectures, and \emph{implicit ensembles}: single models that approximate deep ensembles, usually through parameter sampling \cite{gal2016dropout}.  To construct heterogeneous ensembles,
we divide the 137 CIFAR10 models and 78 ImageNet models from \cref{sec:setup} based on their InD accuracy.
Ensembles are then formed by randomly selecting 4 models from each bin.
This procedure ensures that all ensemble members will have similar accuracy, even though the ensemble members may represent different architectures and training regimens.
Despite their additional diversity, these heterogeneous ensembles do not provide effective robustness, as shown in 
 \cref{sec:app_heter_ensemble}. Finally, we investigate if these results also follow for three implicit ensembling mechanisms: Monte Carlo Dropout \cite{gal2016dropout}, multiple-input-multiple-output (MIMO) \cite{havasi2021training}, and Batch Ensembles \cite{wen2020batchensemble}.  We find that implicit ensembles are also not effectively robust, as depicted in \cref{sec:app_heter_ensemble}.

\subsection{Implications.}
As discussed in \cref{sec:correlated_improvemenets_pp}, ensemble diversity is responsible for improved NLL and Brier score relative to constituent models.
In this sense, ensemble diversity is responsible for improved OOD performance.
However, these OOD improvements exactly follow the deterministic trends predicted by (standard) single models, and thus ensembling multiple diverse predictors does not yield any ``effective robustness'' over what could be achieved by a better performing single model.
Unlike prior research \citep[e.g.][]{ovadia2019can,gustafsson2020evaluating}, these results suggest that ensembles are a tool of convenience for obtaining better OOD performance, but not qualitatively different from single models in this respect.

\section{Discussion}
\label{sec:discussion}

In this work, we rigorously test common intuitions about the benefits of deep ensembles to UQ and robustness, and find these explanations wanting. 
Below, we lay out limitations of our study, summarize our conclusions, and indicate important lines of future work. 

\textbf{Ensembling in the overparametrized regime.} We emphasize that our analysis only focuses on ensembles of neural networks, and does not necessarily apply to ensembling techniques in general (e.g. random forests or gradient boosted decision trees). 
Indeed, we predict that many of our results are direct consequences of the fact that we are ensembling high-capacity ``interpolating" models, which seem to generalize well despite being massively overparametrized \citep{belkin2019reconciling,adlam2020understanding,nakkiran2021deep,hastie2022surprises}. In future work, we will examine the effect of overparametrization directly by replicating these experiments with ensembles of weak learners.

\textbf{Neural network uncertainty quantification.} 
In examining the conditional distributions in \cref{fig:f0}, we see that OOD uncertainty quantification is not directly impacted by ensemble diversity.
These findings show that the role of ensemble diversity in deep ensemble UQ is far more limited than previously hypothesized \citep[e.g.][]{lakshminarayanan2016simple,fort2019deep,gustafsson2020evaluating}.

\textbf{Effective robustness.} Our results in Figure~\ref{fig:ltrend_metrics} show that ensemble diversity does not yield improvements to robustness that cannot be explained by InD performance. This finding is in line with other results demonstrating that effective robustness is very difficult to achieve \cite{andreassen2021evolution}.

\textbf{When should we use deep ensembles?}
Despite our results, we maintain that
ensembling can be viewed as a reliable ``black box'' method of improving neural network performance,
both InD and OOD. 
It is simple (though potentially expensive) to improve upon a model through ensembling, and training a single model that matches the performance of an ensemble is not always straightforward \citep{kondratyuk2020ensembling,lobacheva2020power,wasay2020more}. 
However we caution that deep ensembles are not a panacea for the issues faced by single models.  
In particular, it is dangerous to assume that deep ensembles mitigate the robustness issues of single models in contexts where we can expect dataset shift, or that ensemble diversity provides a reliable baseline for model uncertainty in the absence of ground truth.
Thus, for many practitioners, the choice of using a deep ensemble versus a performance matched single model may ultimately be dictated by practical considerations, such as performance given a pre-determined parameter/FLOP budget for model training and evaluation \citep{kondratyuk2020ensembling,lobacheva2020power,wasay2020more}.
Beyond these practical concerns, we have yet to find evidence for any reason to prefer the use of deep ensembles over an appropriately chosen single model.

\begin{ack}
We thank John Miller for sharing models trained on CIFAR10,
and \citet{taori2020measuring} for making their trained ImageNet models and code open sourced and easy to use.
We would also like to thank Dustin Tran for his insightful comments, and Julien Boussard for helpful discussions on statistical testing. 
TA is supported by NIH training grant 2T32NS064929-11.
EKB is supported by NIH 5T32NS064929-13, NSF 1707398, and Gatsby Charitable Foundation GAT3708.
GP and JPC are supported by the Simons Foundation, McKnight Foundation, Grossman Center for the Statistics of Mind, and Gatsby Charitable Trust.
\end{ack}
\bibliography{citations}
\bibliographystyle{plainnat}
\section*{Checklist}


\begin{enumerate}

\item For all authors...
\begin{enumerate}
  \item Do the main claims made in the abstract and introduction accurately reflect the paper's contributions and scope?
    \answerYes{We relate our claims to two hypotheses in the introduction \cref{sec:intro}, and test each hypothesis in the corresponding results sections \cref{sec:uncertainty,sec:ltrend}.} 
  \item Did you describe the limitations of your work?
    \answerYes{We specify in the discussion \cref{sec:discussion} that our results are limited to neural network ensembles, as opposed to more general ensembles.}
  \item Did you discuss any potential negative societal impacts of your work?
    \answerYes{We do so in \cref{sec:societal_impact}}
  \item Have you read the ethics review guidelines and ensured that your paper conforms to them?
    \answerYes{}
\end{enumerate}

\item If you are including theoretical results...
\begin{enumerate}
  \item Did you state the full set of assumptions of all theoretical results?
    \answerYes{}
        \item Did you include complete proofs of all theoretical results?
    \answerYes{}
\end{enumerate}

\item If you ran experiments...
\begin{enumerate}
  \item Did you include the code, data, and instructions needed to reproduce the main experimental results (either in the supplemental material or as a URL)?
    \answerYes{We provide a link to a repository in the supplemental material section \cref{sec:app_code_data}. This repository contains instructions to reproduce main figures and to download relevant data.  }
  \item Did you specify all the training details (e.g., data splits, hyperparameters, how they were chosen)?
    \answerYes{We provide instructions in the code \cref{sec:app_code_data} which specify internally the data splits and hyperparameters we used. We further specify in \cref{sec:app_training_details} that we chose default hyperparameters as specified in a separate code repo.}
        \item Did you report error bars (e.g., with respect to the random seed after running experiments multiple times)?
    \answerYes{In accordance with checklist guidelines, we report the fact that we ran statistical significance tests for our main results here- in particular, \cref{sec:appconddiv} describes tests for \cref{fig:f0} and related results, \cref{sec:appmmd} describes tests for \cref{fig:perfcomp} and related results, and \cref{sec:additional_ltrend} describes tests for \cref{fig:ltrend_metrics} and related results.}
        \item Did you include the total amount of compute and the type of resources used (e.g., type of GPUs, internal cluster, or cloud provider)?
    \answerYes{We do so in \cref{sec:app_code_data}}
\end{enumerate}

\item If you are using existing assets (e.g., code, data, models) or curating/releasing new assets...
\begin{enumerate}
  \item If your work uses existing assets, did you cite the creators?
    \answerYes{In \cref{sec:setup}, we reference the origin of the models \cite{miller2021accuracy} and \cite{taori2020measuring}, and provide further details in \cref{sec:app_code_data}.}
  \item Did you mention the license of the assets?
    \answerNo{We provide links to publicly released assets with relevant licenses, but do not have a license for models that we were provided by the authors of \cite{miller2021accuracy}.}
  \item Did you include any new assets either in the supplemental material or as a URL?
    \answerNA{We do not provide new assets.}
  \item Did you discuss whether and how consent was obtained from people whose data you're using/curating?
    \answerYes{In our acknowledgements we thank the authors of \cite{miller2021accuracy} for agreeing to share their data with us- all other data is released under a pubic license.}
  \item Did you discuss whether the data you are using/curating contains personally identifiable information or offensive content?
    \answerNA{We do not believe this to apply to our data, which consists of deep network models trained using popular deep learning frameworks on benchmark datasets.}
\end{enumerate}

\item If you used crowdsourcing or conducted research with human subjects...
\begin{enumerate}
  \item Did you include the full text of instructions given to participants and screenshots, if applicable?
    \answerNA{}
  \item Did you describe any potential participant risks, with links to Institutional Review Board (IRB) approvals, if applicable?
    \answerNA{}
  \item Did you include the estimated hourly wage paid to participants and the total amount spent on participant compensation?
    \answerNA{}
\end{enumerate}

\end{enumerate}

\newpage
\appendix
\onecolumn

\section{Societal impact}
\label{sec:societal_impact}

Deep ensembles are popular in many real world applications, and a potential negative impact of our work is to expose flaws in applications reliant upon deep ensembles, especially in adversarial settings like fraud detection (although this may lead to improved systems further on as well).

\section{Code, data and compute}
\label{sec:app_code_data}
\subsection{Code and data}
We provide general directions to reproduce the main results of our paper in the linked directions here: \url{https://github.com/cellistigs/interp_ensembles#readme}. 

These directions reference two repositories, corresponding to two separate branches of our codebase. The ``compare\_performance" branch can be found here: \url{https://github.com/cellistigs/interp_ensembles/tree/compare_performance}. Likewise, the ``imagenet\_pl" branch can be found here: \url{https://github.com/cellistigs/interp_ensembles/tree/imagenet_pl}. 
This code repository, together with the instructions provided above, specify all training and visualization details relevant to our study. 

Finally, we share relevant data as a Zenodo repository: \url{https://zenodo.org/record/6582653#.Yo7R0y-B3fZ}. This data provides the logit outputs from individual models (and some ensembles) on the in and out of distribution data that we consider. These data are referenced in the code above. 

\subsection{Compute}

We ran all CIFAR10 model training on Amazon Web Services (AWS), using the ``p3.2xlarge" instance type with a Tesla V100 GPU. We ran half of ImageNet model training on an internal cluster with GeForce RTX 2080 Ti GPUs, and the other half on AWS with the ``p3.8xlarge" instance type, again with Tesla V100 GPUs. Visualization and statistical testing was run on M1 MacBook Airs, and additionally on  AWS ``p3.2xlarge" and ``p3.8xlarge" instances when additional capacity was required.

We show results for 50 models trained on CIFAR10, and 15 models trained on ImageNet. We estimate that on average, our CIFAR10 models required 3 hours of compute to train, and our ImageNet models required 48 hours to train. Finally, we estimate an additional 8 hours of compute required to run statistical tests and visualize results,  resulting in a total of approximately 878 hours of total compute.  
\section{Decompositions for uncertainty metrics\label{sec:uncertdecomp}}

\subsection{Jensen-Shannon divergence and entropic uncertainty}

If we use Jensen-Shannon divergence (Eq.~\ref{eqn:variance}) as a metric for ensemble diversity \citep{lakshminarayanan2016simple,fort2019deep}, we show ensemble uncertainty can be decomposed as:
\begin{align}
  \overbracket{
    \entropy \left[ y \! \mid \! \bar \vf(\vx) \right]
  }^{\text{ens. uncert.}}
  = \overbracket{
    \jsd_{p(\vf)} \left[ y \! \mid \! \vf(\vx) \right]
  }^{\text{ens. diversity}}
  + \overbracket{
    \E_{p(\vf)} \left[  \entropy \left[ y \! \mid \! \vf(\vx) \right] \right]
  }^{\text{avg. single model uncert.}}.
  \label{eqn:jsd_uncertainty}
\end{align}
where $\entropy [ y \! \mid \! \cdot ]$ represents the entropy of a categorical distribution parameterized by $(\cdot)$.

Furthermore, $ \jsd_{p(\vf)} \left[ y \! \mid \! \vf(\vx) \right] = \frac{1}{M}\sum_{m=1}^{M}KL[y \! \mid \!  \vf(\vx) \| y \! \mid \! \bar \vf(\vx)  ]$, the average KL divergence between individual model predictions and the ensemble prediction. 

We write:
\begin{align*}
    \entropy \left[ y \! \mid \! \bar \vf(\vx) \right] &= - \frac{1}{C}\sum_{i}p(y_i\mid \bar{\vf})\log(p(y_i\mid \bar{\vf})) \\ 
    & = -\frac{1}{C}\sum_{i}\frac{1}{M}\sum_j p(y_i\mid \vf_j)\log(p(y_i\mid \bar{\vf})) \\ 
    & = -\frac{1}{M}\sum_{j}\frac{1}{C}\sum_i p(y_i\mid \vf_j)\log(p(y_i\mid \bar{\vf})) \\
    & = -\frac{1}{M}\sum_{j}\frac{1}{C}\sum_i p(y_i\mid \vf_j)\left[\log(p(y_i\mid \bar{\vf})) - \log(p(y_i\mid \vf_j)) + \log(p(y_i\mid \vf_j)) \right] \\ 
    & = -\frac{1}{M}\sum_{j}\frac{1}{C}\sum_i p(y_i\mid \vf_j)\left[\log[\frac{p(y_i\mid \bar{\vf})}{p(y_i\mid \vf_j)}] + \log(p(y_i\mid \vf_j)) \right] \\ 
    & = -\frac{1}{M}\sum_{j}\frac{1}{C}\sum_i p(y_i\mid \vf_j)\left[\log(\frac{p(y_i\mid \bar{\vf})}{p(y_i\mid \vf_j})\right] + -\frac{1}{M}\sum_{j}\frac{1}{C}\sum_i p(y_i\mid \vf_j) \log p(y_i\mid \vf_j)) \\ 
    & = \jsd_{p(\vf)} \left[ y \! \mid \! \vf(\vx) \right] + \E_{p(\vf)} \left[  \entropy \left[ y \! \mid \! \vf(\vx) \right] \right]
\end{align*}

\subsection{Variance and quadratic uncertainty}

As in the main text, we provide a decomposition for a quadratic notion of uncertainty as:
\begin{align}
     U \left( \bar \vf(\vx) \right) = \Var_{p(\vf)} \left[ \vf(\vx) \right] + \E_{p(\vf)} \left[  U\left( \vf(\vx) \right) \right] 
\end{align}
where $U( \vf(\vx) )$ is a quadratic notion of uncertainty:
\[
    U \left( \vf(\vx) \right) \triangleq 1 - \sum_{i=1}^C \bigl[ p(y = i \mid \vf(\vx)) \bigr]^2.
\]

And variance is defined as: 
\[
\Var_{p(\vf)} \left[ \vf(\vx) \right] = {\textstyle \sum_{i=1}^C } \Var_{p(\vf)} \left[ f^{(i)} (\vx) \right]
\]
Then, the ensemble uncertainty can be decomposed as follows: 
\begin{align*}
    U \left( \bar{\vf}(\vx) \right) &= 1 - \sum_{i=1}^C \bigl[ p(y = i \mid \bar{\vf}(\vx)) \bigr]^2 \\
    & = 1 - \E_{p(\vf)}[\sum_{i=1}^C \bigl[ p(y = i \mid \vf(\vx)) \bigr]]^2 + \E_{p(\vf)}[\sum_{i=1}^C \bigl[ p(y = i \mid \vf(\vx)) \bigr]]^2 - \sum_{i=1}^C \bigl[ p(y = i \mid \bar{\vf}(\vx)) \bigr]^2 \\
    & = \E_{p(\vf)}[1 - \sum_{i=1}^C \bigl[ p(y = i \mid \vf(\vx)) \bigr]]^2 + \sum_{i=1}^C\E_{p(\vf)}[\bigl[ p(y = i \mid \vf(\vx)) \bigr]]^2- \bigl[ \E_{p(\vf)} [p(y = i \mid \vf(\vx))] \bigr]^2 \\
    & = \E_{p(\vf)}[ U \left( \vf(\vx) \right)] + \Var_{p(\vf)} \left[ \vf(\vx) \right]
\end{align*}

\section{Brier score Jensen gap}
\label{sec:app_brier_score_uncertainty_decomposition}
We consider the Brier Score of a single model:
\begin{align}
    \E_\vf \left[ B_p(\vf_i) \right] &= \E_{p(\vx, y)} \E_\vf \left[ \Vert \vf_i(\vx) - \vone_y \Vert_2^2 \right]
    \label{eqn:avg_brier} \\
    &= \E_{p(\vx, y)} \left[
        \E_\vf \left[ \Vert \vf_i(\vx) \Vert_2^2 \right]
        + 2 \bar \vf_(\vx)^\top \vone_y
        + 1
    \right]
    \nonumber
\end{align}
and the Brier score of the ensemble:
\begin{align}
    B_p(\bar \vf) &= \E_{p(\vx, y)} \left[ \Vert \bar \vf(\vx) - \vone_y \Vert_2^2 \right]
    \label{eqn:ens_brier} \\
    &= \E_{p(\vx, y)} \left[
        \Vert \bar \vf(\vx) \Vert_2^2
        + 2 \bar \vf(\vx)^\top \vone_y
        + 1
    \right]
    \nonumber
\end{align}
Note that \cref{eqn:avg_brier} and \cref{eqn:ens_brier} only differ by a single term:
\begin{align*}
    B_p(\bar \vf) &= \E_\vf \left[ B_p(\vf) \right]
    + \E_{p(\vx)} \left[ \Vert \bar \vf(\vx) \Vert_2^2  \right]
    \\ &\phantom{=}
    - \E_{p(\vx)} \E_\vf \left[ \Vert \vf(\vx) \Vert_2^2 \right] \\ 
    & = \E_\vf \left[ B_p(\vf) \right] - \E_{p(\vx)}[\Vert \E_{\vf}\vf(\vx) \Vert_2^2- \Vert \bar \vf(\vx) \Vert_2^2 ] \\
    & = \E_\vf \left[ B_p(\vf) \right] - \E_{p(\vx)}[\E_{\vf}[\Vert \vf(\vx) \Vert_2^2]- \Vert \E_{\vf}[\vf(\vx)] \Vert_2^2 ] \\
    & = \E_\vf \left[ B_p(\vf) \right] - \E_{p(\vx)}[\Var_{p(\vf)} \left[ \vf(\vx) \right]]
\end{align*}

Where $\Var_{p(\vf)} \left[ \vf(\vx) \right]$ is:

$$\Var_{p(\vf)} \left[ \vf(\vx) \right]= {\textstyle \sum_{j=1}^C } \Var_{p(\vf)} \left[ f^{(j)} (\vx) \right]$$.

We note that this relation holds at the level of individual data points as well.

\section{Expected behavior of Bayesian model average on InD/OOD uncertainty quantification}
\label{sec:gp_uncertainty}

As a motivating example, we consider uncertainty quantification on InD and OOD data using a Bayesian model average, and relate our findings back to the implications presented in \cref{sec:implications_epis_alea}. An ideal Bayesian model average should express higher posterior variance on OOD data than InD data, even after controlling for other sources of uncertainty.
To demonstrate this desired behavior in practice, we consider Gaussian processes, a class of models well regarded for its uncertainty quantification capabilities \cite{williams2006gaussian}.
The Gaussian process model $f(\cdot)$ is defined by the following generative process:
\begin{equation}
    \begin{split}
    p(f(\cdot)) &= \mathcal{GP}, \\
    p(y \mid f(x)) &= \mathcal{N}(0, \sigma^2(x))
    \end{split}
    \label{eqn:gp_process}
\end{equation}
where $\sigma^2(x)$ is a heteroskedastic noise function defined as $\sigma^2(x) = \sin^2(x) + 0.01$.
After conditioning on training data $\mathcal D$, the BMA at a test point $x$ is given by:
\begin{equation}
    p(y \mid x, \mathcal D) = \mathcal N( \mu_{f \mid \mathcal D}(x), \mathrm{Var}_{f \mid \mathcal D}(x) + \sigma^2(x) ),
    \label{eqn:gp_posterior}
\end{equation}
where $\mu_{\mid \mathcal D}(\cdot)$ and $\mathrm{Var}_{f \mid \mathcal D}(\cdot)$ are the posterior predictive GP mean and variance, respectively, which can both be computed in closed form.
(See \cite{williams2006gaussian} for closed-form expressions for these two functions).
Crucially, the predictive variance in \cref{eqn:gp_posterior} is a uncertainty estimate that decomposes into epistemic and aleatoric components:
the {\bf posterior variance} term ($\mathrm{Var}_{f \mid \mathcal D}(\cdot)$) 
and the {\bf likelihood variance} term ($\sigma^2(x)$), respectively. 

 \begin{figure}[t!]
 \centering
 \includegraphics[width=\linewidth]{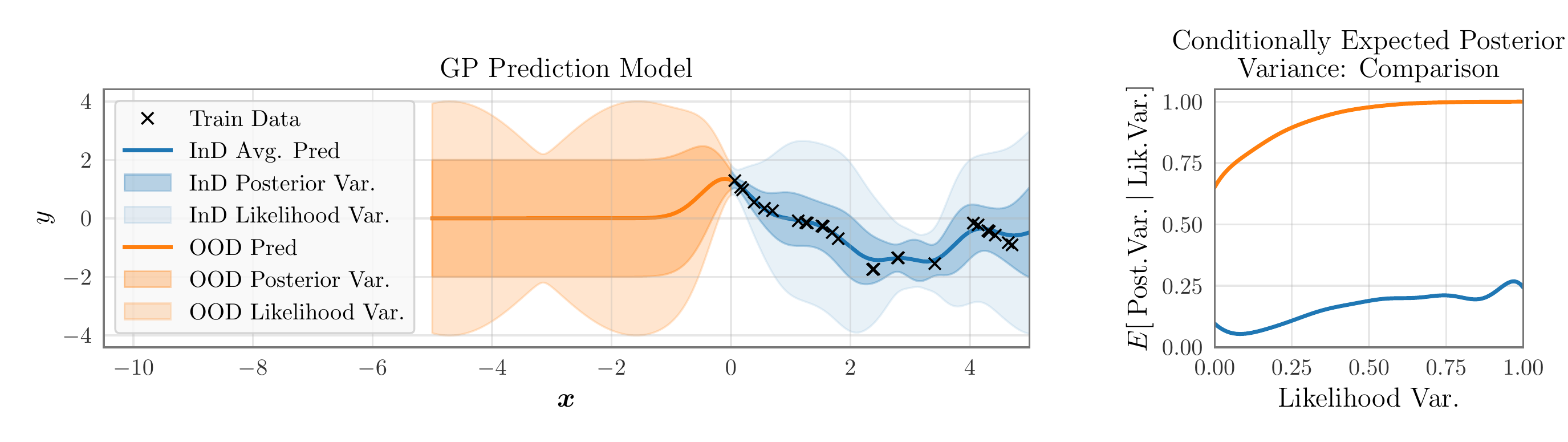}
 \caption{
     Example of a model where OOD predictions have higher posterior variance, even after controlling for other sources of uncertainty.
     {\bf Left:} The predictive uncertainty expressed by a Gaussian process model on a toy regression dataset.
     OOD data (orange) express higher posterior variance than InD data (blue).
     {\bf Right:} The expected posterior variance ) conditioned on a prediction's likelihood variance is also significantly larger for OOD data.
 }
 \label{fig:gp}
 \end{figure}

In \cref{fig:gp} (left), we generate a one-dimensional dataset by drawing 25 random data points over $x \in [0, 5]$ using the generative process defined in \cref{eqn:gp_process}.\footnote{
    In all experiments, the prior GP model has zero mean and a RBF covariance function with a lengthscale of 1. 
}
After fitting a GP model to these data, we compute the predictive posterior over the range $x \in [-5, 5]$.
The points in $[0, 5]$ represent InD data---as they share the same domain as the training data---while the points in $[-5, 0]$ (orange) represent OOD data.
In \cref{fig:gp} (right), we observe that OOD predictions have much higher expected posterior variance, even after conditioning on a prediction's likelihood uncertainty.
Note that this is in stark contrast to the analogous deep ensemble results in \cref{sec:uncertainty}, where there is little to no conditional difference between OOD and InD predictions.
\section{Quantifying conditional diversity\label{sec:appconddiv}}

In this section, we provide additional experimental details for the results in \cref{fig:f0}, and extend to other datasets and measures of ensemble diversity. We also introduce quantifications and signficance tests to validate the stability of our conclusions across many combinations of OOD dataset and model. 

\subsection{Marginal distribution of average single model uncertainty \label{appx:avg_plots}}

We end \cref{sec:diversity_exp} with the surprising conclusion that any changes to ensemble UQ between InD and OOD data must come from changes in the distribution of average single model uncertainty, $p(\mathbb{E}(U(f(x))$. Here we confirm empirically that this distribution does shift towards higher uncertainty on OOD data, for the same models that we present in \cref{sec:diversity_exp}. This shift drives any changes in ensemble diversity that we observe in practice. 

\begin{figure*}[!htb]
\centering
\subfloat{\includegraphics[width=0.5\textwidth]{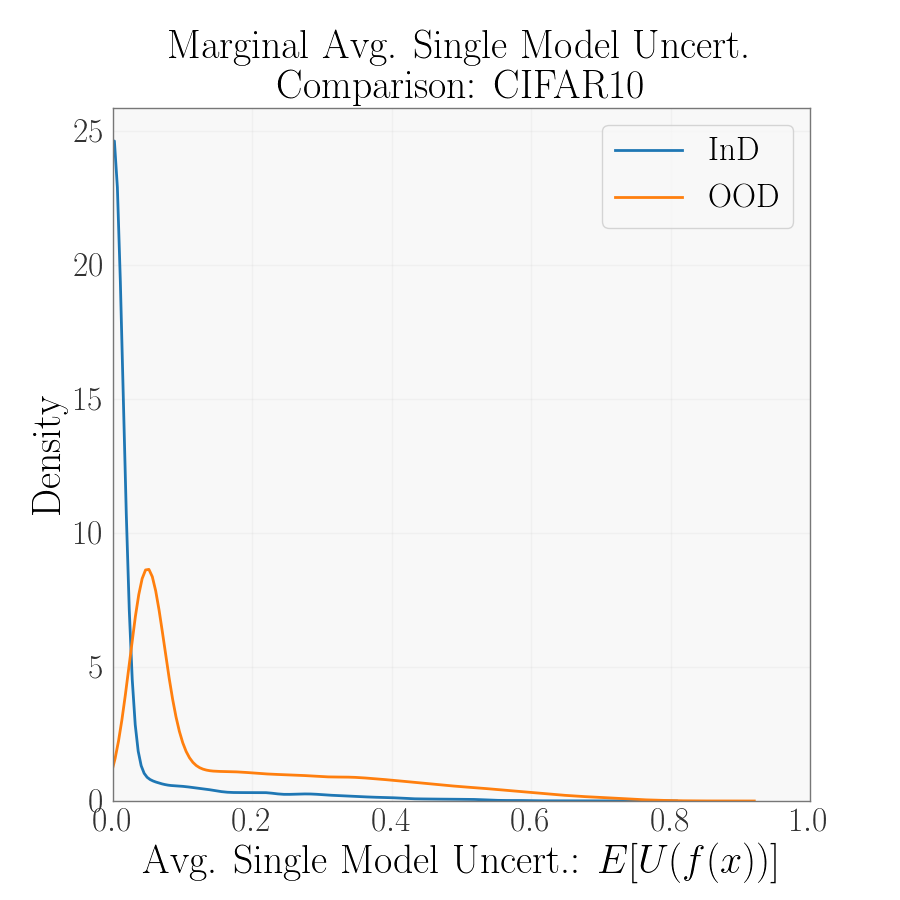}}
\subfloat{\includegraphics[width=0.5\textwidth]{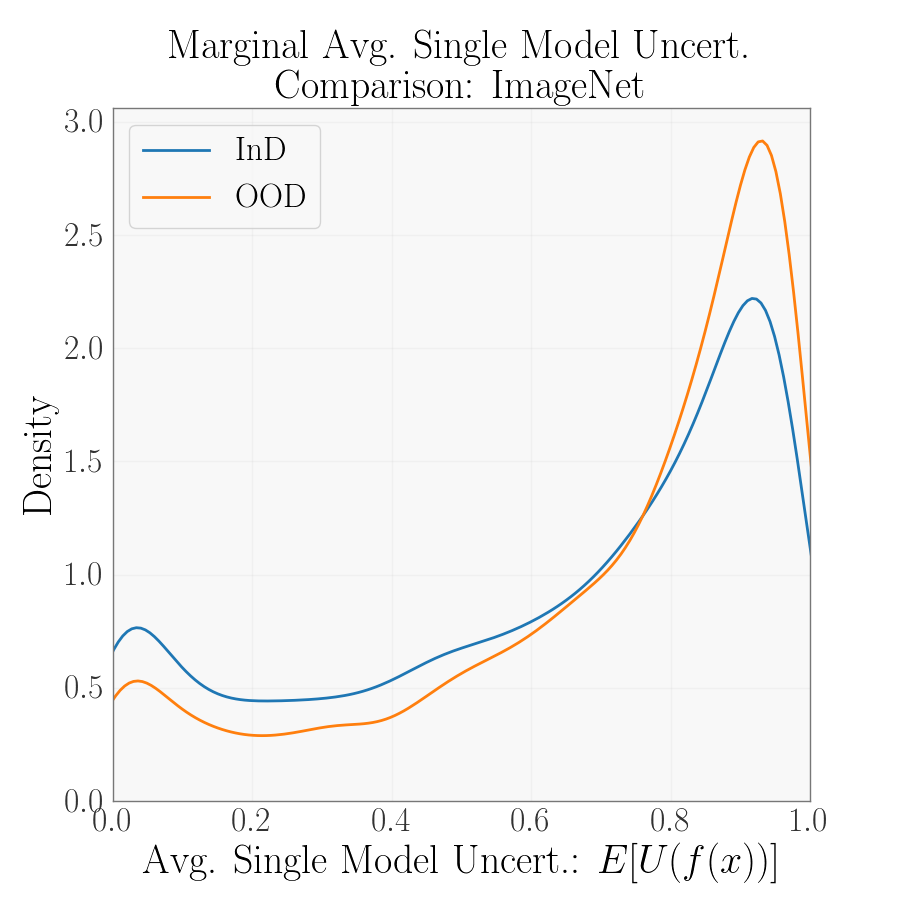}}
\caption{Distributions of average single model uncertainty for the WideResNet 28-10 ensembles trained on CIFAR10 (left) and the AlexNet ensembles (right), as in \cref{fig:f0}. InD and OOD test datasets are CIFAR10 and CINIC10 for the left panel, and ImageNet and ImageNet V2 for the right. }
\label{fig:sigtest_Var}
\end{figure*}

\subsection{Generating conditional distributions and conditional expectations}
In order to depict conditional variance distributions, we fit kernel density estimates to the joint distribution of ensemble diversity and average single model uncertainty for all evaluation datasets. We generated KDEs with the bandwidth suggested by Scott's Rule, and approximate conditional distributions by dividing each column of our KDE grid by the average value. 

To validate comparisons between conditional distributions more precisely, we estimate the conditional expectation $\E[\text{Diversity}\mid\text{Avg}]$ by fitting a Kernel Ridge Regression model to these data, giving the best fit curve to predict values of ensemble diversity from a given value of average single model uncertainty. We used a Gaussian kernel, with bandwidth identical to what was used to generate KDE plots. 

Strictly to ease visualization, we generated conditional expectation estimates for CINIC10 with a randomly subsampled set of 10000 points when fitting Kernel Ridge Regression. We account for any potential bias this may introduce in our statistical quantifications below.

\subsection{Visualizations for other datasets and metrics}

\begin{figure*}[!htb]
\centering
\captionsetup{justification=centering}

\includegraphics[width=1\textwidth]{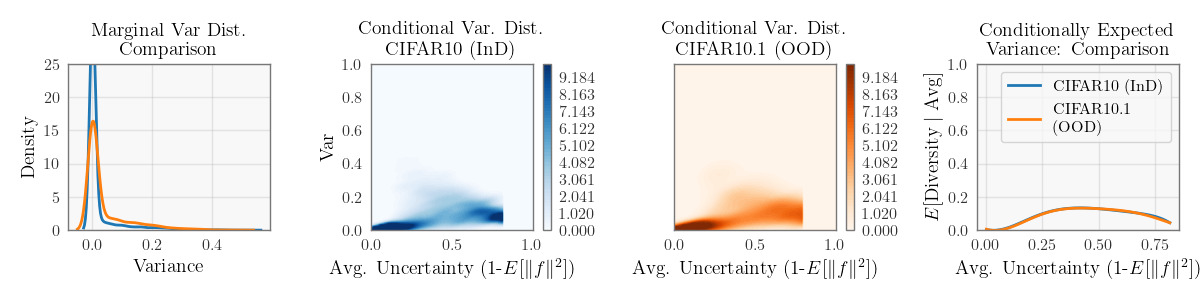}]
\includegraphics[width=1\textwidth]{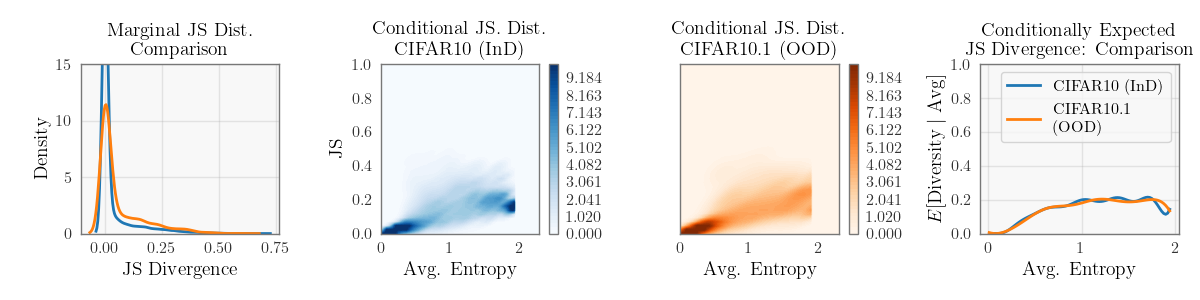}]
\includegraphics[width=1\textwidth]{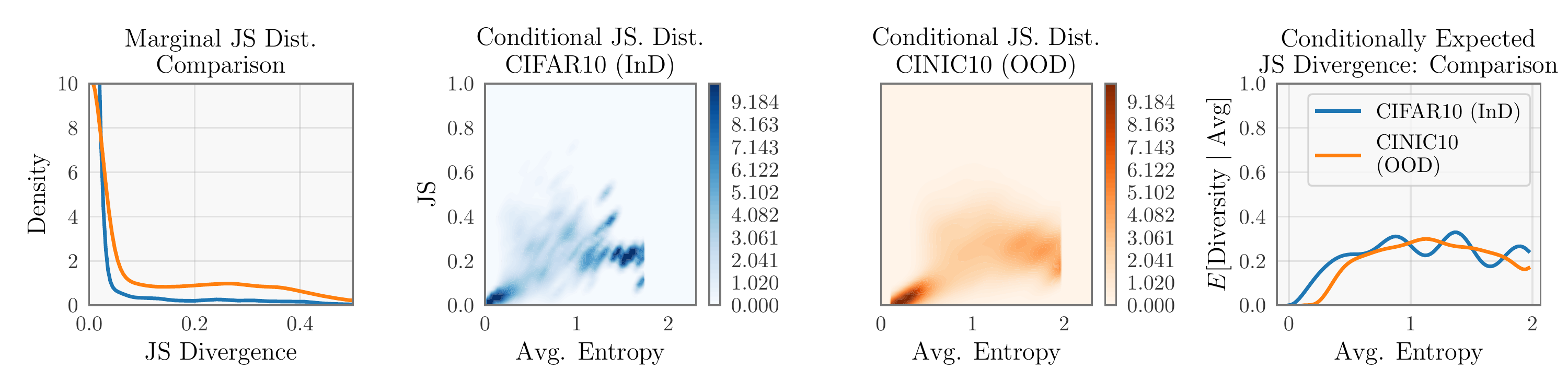}]
\includegraphics[width=1\textwidth]{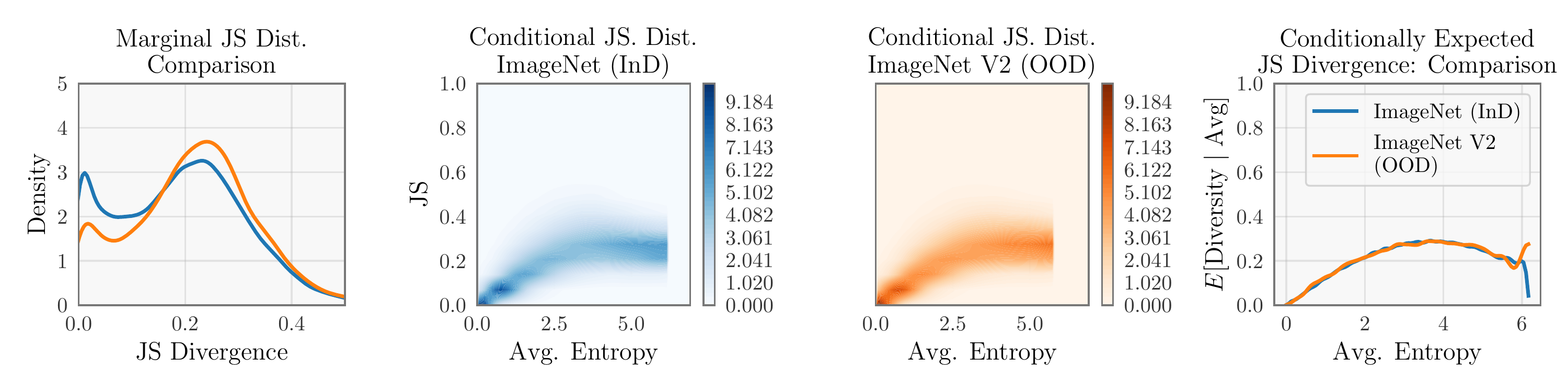}]
\vspace{-0.15in}
\caption{The top panels illustrates the InD vs OOD Variance for Cifar 10 vs Cifar10.1 with an ensemble of 4 VGG-11 networks. The bottom 3 panels illustrate the JS divergence  on InD (Blue) and OOD (orange) data for CIFAR10 vs CIFAR10.1 (VGG-11), CIFAR10 vs CINIC10 (WideResNet-28-10) and ImageNet vs ImageNetV2 (AlexNet).\\  Conventions and conclusions as in Figure~\ref{fig:f0}.}
\label{fig:f0_appendix}
\end{figure*}

Figure \ref{fig:f0_appendix} first shows the variance analysis that we conducted extended to CIFAR10/CIFAR10.1, estimated with an ensemble of 5 VGG 11 networks. In the rows below, we show all analogous conclusions for Jensen Shannon Divergence as a measure of ensemble diversity, instead of variance for the same models (ensembles of $M=4$ VGG-11, WideResNet28-10, and AlexNet models for CIFAR10.1, CINIC10 and ImageNet V2 respectively). 
Across all datasets, we observe that the same trends hold as reported in \cref{fig:f0}. Namely, ensemble diversity is higher on OOD data than InD data, but that the corresponding conditional distributions are not distinguishable.

\subsection{Large scale quantification and statistical tests}

In order to scale these analyses further, we devised a test statistic to directly compare the conditional expected diversity measures of InD and OOD data. Given conditional expectations for InD and OOD data, consider the following statistic: 
\begin{align*}
    d(InD,OOD) = \int d\text{Avg} \frac{\E_{OOD}[\text{Diversity}\mid\text{Avg}] -\E_{InD}[\text{Diversity}\mid\text{Avg}]}{\E_{InD}[\text{Diversity}\mid\text{Avg}]}
\end{align*}

Intuitively, this statistic measures the percentage change in area under the conditional expectation curve when we consider an OOD conditional expectation instead of a corresponding InD conditional expectation.

\begin{figure*}[!htb]
\centering
\includegraphics[width=1\textwidth]{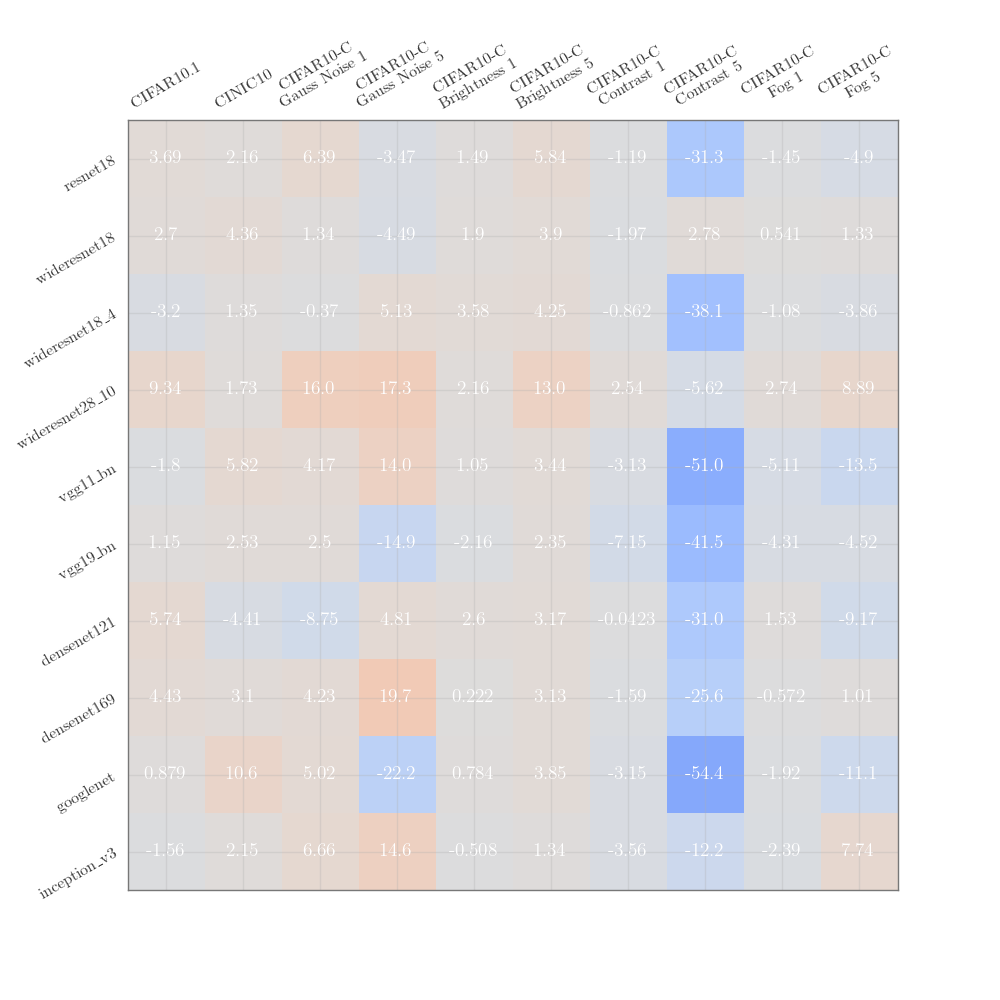}
\caption{Percent Increase (OOD over InD) for Variance  Decomposition}
\label{fig:percentincrease_Var_cifar}
\end{figure*}

\begin{figure*}[!htb]
\centering
\includegraphics[width=1\textwidth]{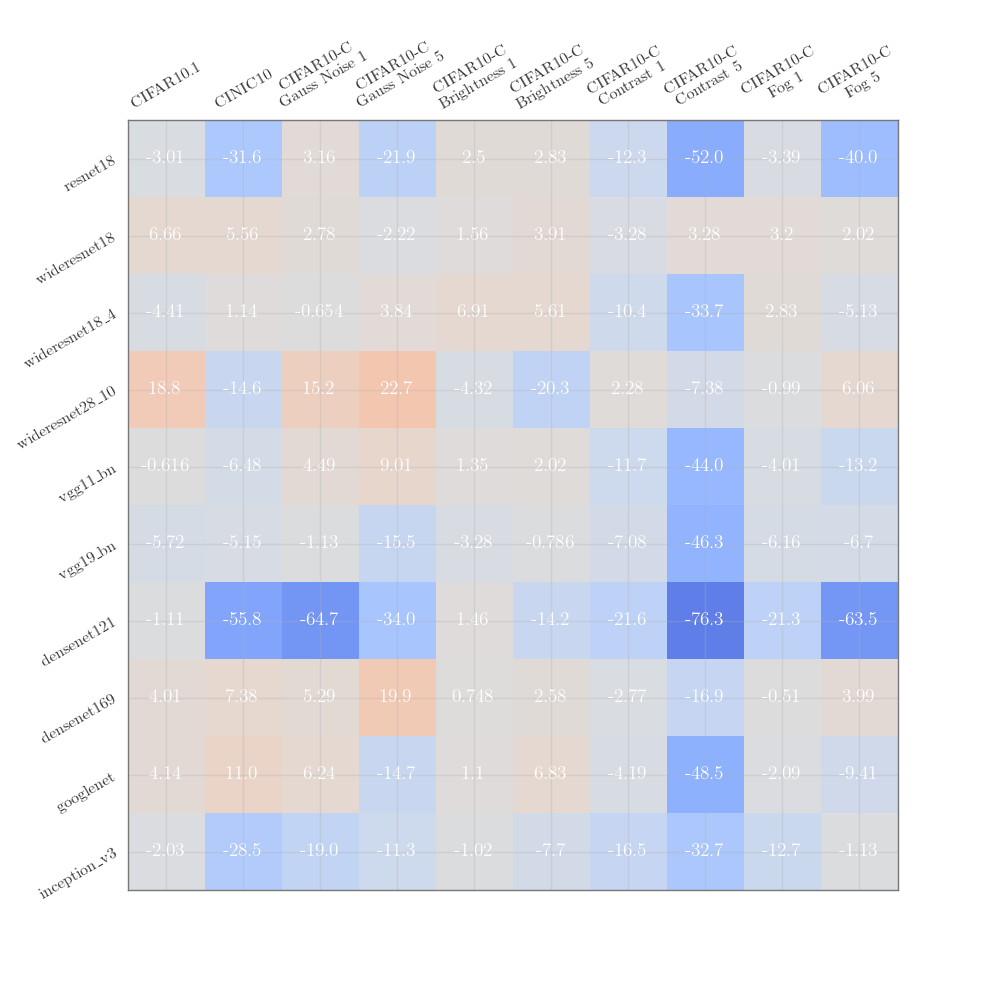}
\caption{Percent Increase (OOD over InD) Jensen Shannon Decomposition}
\label{fig:percentincrease_JS_cifar}
\end{figure*}

We approximated this percentage increase in expected conditional diversity as sum of pointwise differences between InD and OOD, divided by the sum of the InD curve, and report results for all model and dataset pairs that we tested in \cref{fig:percentincrease_Var_cifar},\cref{fig:percentincrease_JS_cifar}. Altogether, we see that in most cases, the percentage increases in area under the OOD curve are very small (for reference, the main text examples demonstrate changes on the order of $\sim 1\%$.) %
Although there are few sporadic cases where certain datasets demonstrate sizeable increases in our statistic on OOD data (consider variance for  DenseNet 169 on CIFAR10-C Gaussian Noise, Severity Level 5), we note that these trends are inconsistent across individual models and datasets, limiting practical use of differences in OOD estimation. Furthermore, we note that our results on natural corruptions (leftmost two columns) are far more consistent than our results on synthetic corruptions (all others). In line with previous work \cite{taori2020measuring}, we prioritize results on natural corruptions in reporting our results.

Next, we performed Monte Carlo permutation tests to quantify the significance of the statistics that we observed: 
\begin{itemize}
    \item For each model and dataset upon which we computed a statistic, we first aggregated all datapoints from in and out of distribution model evaluations, and randomly permuted the order of these samples, generating a surrogate sample. 
    \item We then refit Kernel Ridge Regression to the surrogate sample, and calculated the $d$ statistic that resulted. 
    \item We calculated if the computed $d$ statistic was greater than or less than what we observed on our original sample.
    \item We repeated this process for a total of 100 surrogate samples. 
\end{itemize}

From this process, we can treat the proportion of surrogate samples that exceeded the value of our true test statistic as a p value for the null hypothesis that the d statistic we calculated measures a significant difference between our two original samples (and in particular, that the conditional expectation of ensemble diversity on OOD data is significantly greater than that of ensemble diversity in InD data.)

In order to compute kernel ridge regression efficiently, we used GPytorch \cite{gardner2018gpytorch} with kernel partitioning to refit models many times on a GPU. This process allowed us to compute statistics on the entire CINIC10 evaluation set, alleviating all possibilities for error in visualization due to subsampling. 

\begin{figure*}[!htb]
\centering
\includegraphics[width=1\textwidth]{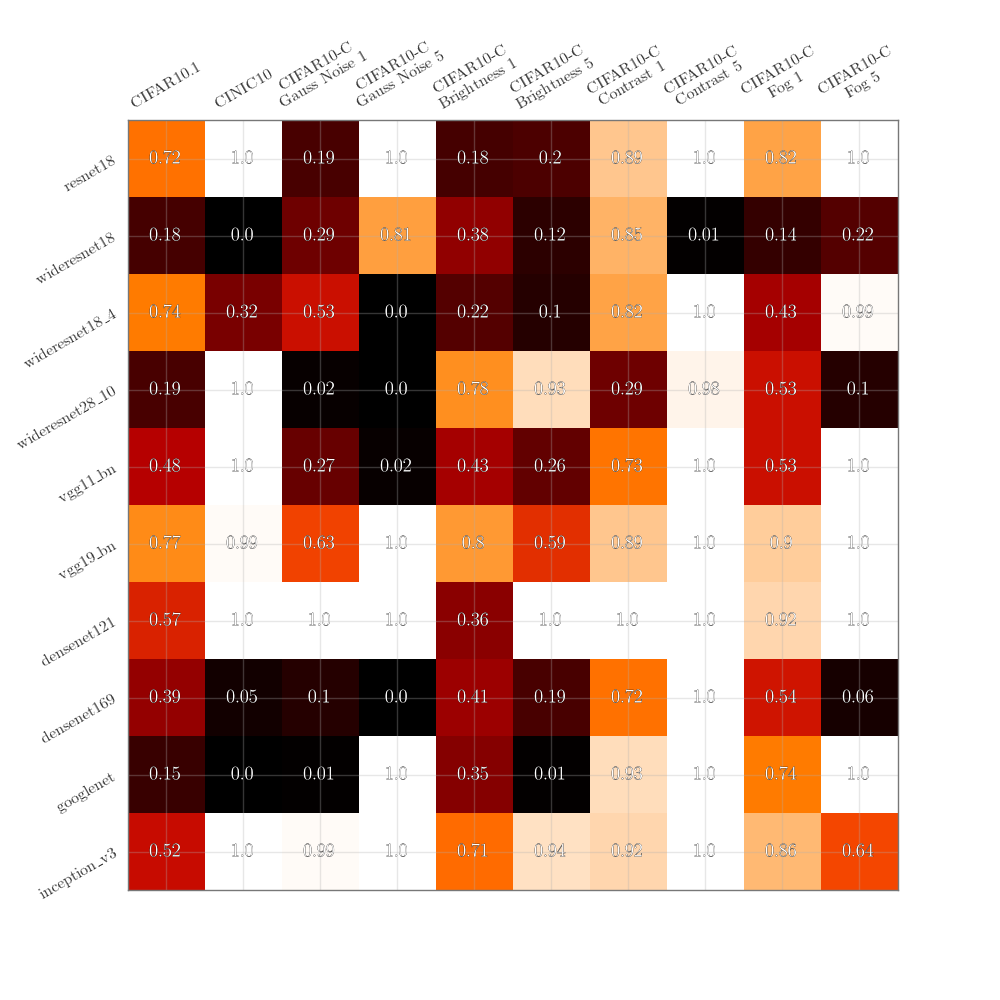}
\caption{P values of (OOD over InD) difference for Jensen Shannon Decomposition}
\label{fig:sigtest_JS}
\end{figure*}

In \cref{fig:sigtest_Var} and \cref{fig:sigtest_JS}, we report the estimated p values from this process. Our main goal is to communicate that in many cases, we found that the differences between conditional expectations for in and out of distribution data were almost certainly not significant, regardless of their absolute magnitude.

Finally, we show percentage increases for Imagenet on analogous $M=5$ ensembles of AlexNet, ResNet 50, and ResNet 101 models \cref{fig:sigtest_Var}, \cref{fig:sigtest_JS}- on ImageNet V2, we once again fail to see any considerable increase on the conditional distributions of OOD data relative to InD data, regardless of metric.

\begin{figure*}[!htb]
\centering
\subfloat{\includegraphics[width=1\textwidth]{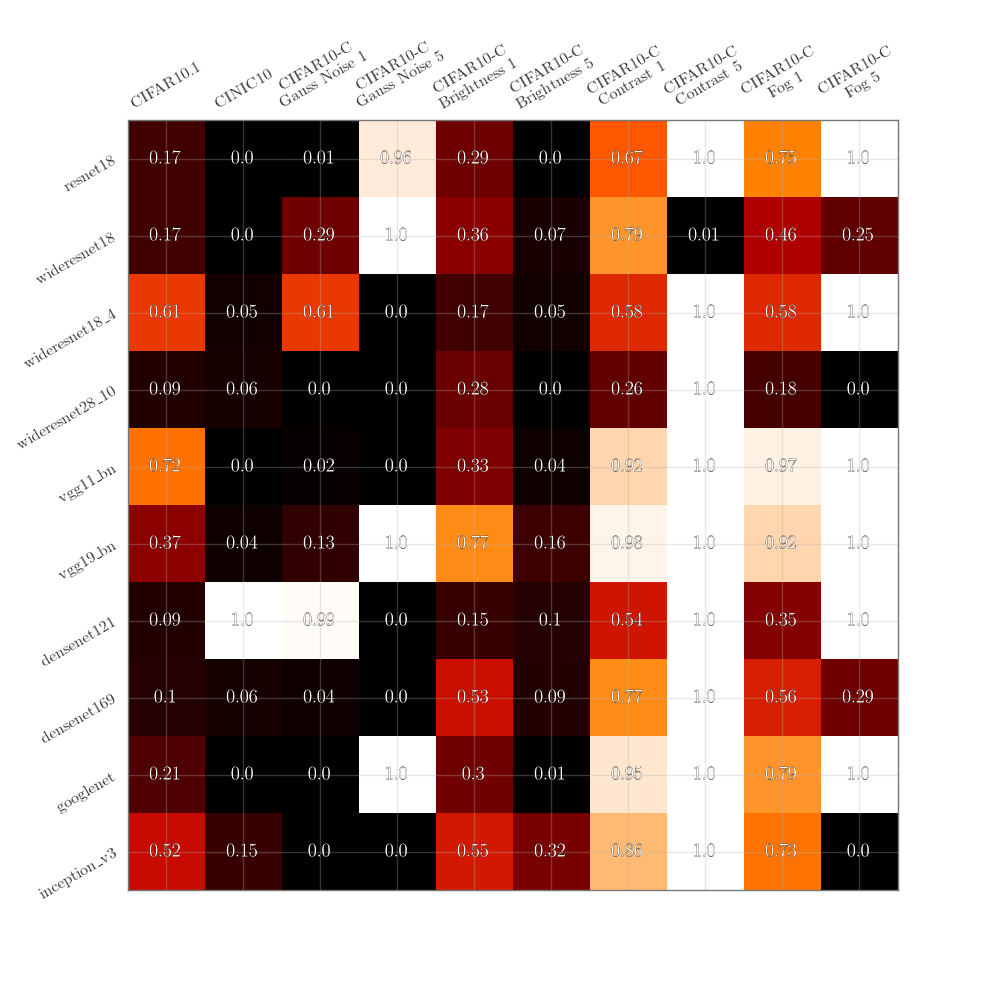}}
\caption{P values of (OOD over InD) difference for Variance Decomposition}
\label{fig:sigtest_Var}
\end{figure*}

\clearpage

We can replicate the finding that differences between in and out of distribution test sets are quite small in the ImageNet dataset as well: 

\begin{figure*}[!htb]
\centering
\subfloat{\includegraphics[width=0.3\textwidth]{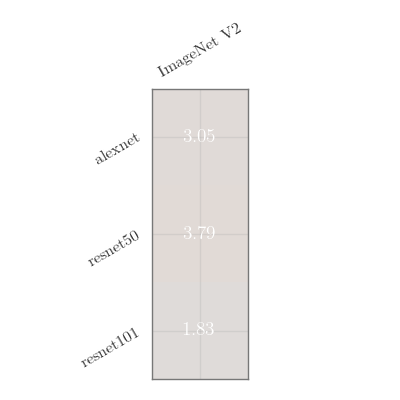}}
\subfloat{\includegraphics[width=0.3\textwidth]{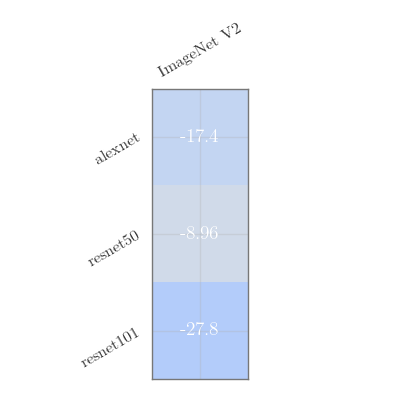}}
\caption{Percent Increase (OOD over InD) Variance Decomposition (left) and JS Divergence (right)}
\label{fig:percentincrease_imagenet}
\end{figure*}

\clearpage 
\section{Details of models for robustness experiments}
\label{sec:app_training_details}
We followed many of the same experimental procedures as \cite{miller2021accuracy} in order to generate ensembles for our experiments. We denote four main groups of models below: 
\subsection{CIFAR10 models trained from scratch}
We trained 10 different classes of models on CIFAR10, noted below. We used implementations from \url{https://github.com/huyvnphan/PyTorch_CIFAR10} in order to train convolutional models adapted for CIFAR10 data sizes, with default hyperparameters, and manually extended existing implementations in this repo to create a WideResNet 18 with width 4. 
\begin{itemize}
    \item ResNet 18 \cite{he2016deep}
    \item WideResNet 18-2, 18-4, 28-10 \cite{zagoruyko2016wide}
    \item GoogleNet, Inception v3 \cite{szegedy2015going}
    \item VGG with 11 and 19 layers \cite{simonyan2014very}
    \item DenseNet 121 and 169 \cite{huang2017densely}
\end{itemize}

We trained five independent instances of each of these architectures with random seeds for 100 epochs each (see code repo defaults for other hyper parameters.)

\subsection{CIFAR10 pretrained ensembles}
We use the models trained by \citet{miller2021accuracy}, and we thank the authors for graciously sharing these results with us.

\subsection{ImageNet models trained from scratch}
We additionally trained two sets of ensembles from scratch on the ImageNet dataset. In particular, we trained 5 model ensembles of AlexNet and ResNet 101 models using implementations available at \url{https://pytorch.org/vision/stable/models.html} for 90 epochs each. 

\subsection{Imagenet pretrained models}

We use 5 of the ResNet50 models trained by \cite{ashukha2020pitfalls} and the standard 78 trained models provided by \citet{taori2020measuring}. 

\section{Additional generalization trend results}
\label{sec:additional_ltrend}

 In this section, we report test statistics for the results we show in \cref{fig:ltrend_metrics}, and we extend the results from \cref{fig:ltrend_metrics} to additional OOD datasets, namely CIFAR10.1 and ImageNet-C \cite{hendrycks2018benchmarking}, illustrating generalization trends for ensembles and individual models for various distortions at different intensity levels.  The results in this section show that for high intensity distortions, single models can break away from a well defined linear trend, as reported in \cite{miller2021accuracy}. However, even at the highest distortion levels, the generalization performance for ensembles and individual models heavily overlap, suggesting the lack of effective robustness demonstrated by deep ensembles is not dependent upon the same phenomena that generate strong trends in single models to begin with. 
 
\subsection{Test statistics for generalization performance trends}
\label{sec:ltrend_r2_tables}
In each table we report the  regression coefficient (Coefficient), the standard error (Std. error) t-statistic, p-value and $R^2$ to reject the null hypothesis that there is no relation between InD and OOD performance for the different metrics considered (left column). The last column indicates the number of models (markers) for each model class depicted in  Fig~\ref{fig:ltrend_metrics}.

Note that we do not apply logit scaling to our axes as in \citep{taori2020measuring}, which was found to increase the fit of linear trend lines. Furthermore, we do not consider non-linear parametrizations of NLL, which could potentially improve the quantification of overlap between single models and ensembles. We consider such parameterizations to be beyond the scope of this work.

\begin{table*}[hbt!]
\centering{
\caption{\textbf{ $R^2$ for InD vs OOD generalization trend fits for different metrics}: CIFAR10 vs CINIC10 in \cref{fig:ltrend_cinic10}.}
\resizebox{.9\textwidth}{!}{%
\begin{tabular}{llrrrrrc}
\toprule
      &          &  Coefficient &  Std. error &  t-statistic &  p-value &    R\textasciicircum 2 &  Number of models \\
Metric & Type &              &             &              &          &        &                   \\
\midrule
\multirow{3}{*}{0-1 Error} & All &        0.038 &       0.002 &       18.981 &      0.0 &  0.853 &               434 \\
      & Single Model &        0.029 &       0.006 &        5.038 &      0.0 &  0.883 &                54 \\
      & Ensemble &        0.039 &       0.002 &       18.349 &      0.0 &  0.848 &               380 \\
\cline{1-8}
\multirow{3}{*}{NLL} & All &        0.116 &       0.006 &       18.285 &      0.0 &  0.894 &               434 \\
      & Single Model &        0.120 &       0.022 &        5.511 &      0.0 &  0.864 &                54 \\
      & Ensemble &        0.116 &       0.007 &       17.559 &      0.0 &  0.896 &               380 \\
\cline{1-8}
\multirow{3}{*}{Brier} & All &        0.051 &       0.003 &       17.415 &      0.0 &  0.876 &               434 \\
      & Single Model &        0.042 &       0.009 &        4.754 &      0.0 &  0.890 &                54 \\
      & Ensemble &        0.052 &       0.003 &       16.754 &      0.0 &  0.873 &               380 \\
\cline{1-8}
\multirow{3}{*}{rESCE} & All &        0.009 &       0.002 &        4.712 &      0.0 &  0.791 &               434 \\
      & Single Model &        0.026 &       0.007 &        3.755 &      0.0 &  0.632 &                54 \\
      & Ensemble &        0.007 &       0.002 &        3.860 &      0.0 &  0.801 &               380 \\
\bottomrule
\end{tabular}
}

}
\label{tab:r2_cinic10}
\end{table*}

\begin{table*}[!ht]
\centering{
\caption{\textbf{ $R^2$ for InD vs OOD generalization trend fits for different metrics}: ImageNet vs ImageNetV2 in \cref{fig:ltrend_imagenetv2}.}
\resizebox{.9\textwidth}{!}{%
\begin{tabular}{llrrrrrc}
\toprule
      &          &  Coefficient &  Std. error &  t-statistic &  p-value &    R\textasciicircum 2 &  Number of models \\
Metric & Type &              &             &              &          &        &                   \\
\midrule
\multirow{3}{*}{0-1 Error} & All &        0.102 &       0.001 &       89.935 &      0.0 &  0.995 &               367 \\
      & Single Model &        0.105 &       0.002 &       43.326 &      0.0 &  0.994 &                93 \\
      & Ensemble &        0.101 &       0.001 &       78.643 &      0.0 &  0.995 &               274 \\
\cline{1-8}
\multirow{3}{*}{NLL} & All &        0.432 &       0.008 &       54.749 &      0.0 &  0.989 &               367 \\
      & Single Model &        0.443 &       0.018 &       24.091 &      0.0 &  0.984 &                93 \\
      & Ensemble &        0.428 &       0.009 &       49.622 &      0.0 &  0.991 &               274 \\
\cline{1-8}
\multirow{3}{*}{Brier} & All &        0.156 &       0.002 &       77.827 &      0.0 &  0.989 &               367 \\
      & Single Model &        0.159 &       0.005 &       34.540 &      0.0 &  0.985 &                93 \\
      & Ensemble &        0.156 &       0.002 &       69.984 &      0.0 &  0.991 &               274 \\
\cline{1-8}
\multirow{3}{*}{rESCE} & All &        0.060 &       0.003 &       19.723 &      0.0 &  0.111 &               367 \\
      & Single Model &        0.067 &       0.006 &       10.871 &      0.0 &  0.090 &                93 \\
      & Ensemble &        0.058 &       0.004 &       16.342 &      0.0 &  0.113 &               274 \\
\bottomrule
\end{tabular}%
}

}
\label{tab:r2_imagenetv2}
\end{table*}

\begin{table*}[htb!]
\centering{
\caption{\textbf{ $R^2$ for InD vs OOD generalization trend fits for different metrics}: CIFAR10 vs CIFAR10.1 in \cref{fig:ltrend_cifar10.1}}
\resizebox{.9\textwidth}{!}{%
\begin{tabular}{llrrrrrc}
\toprule
      &          &  Coefficient &  Std. error &  t-statistic &  p-value &    R\textasciicircum 2 &  Number of models \\
Metric & Type &              &             &              &          &        &                   \\
\midrule
\multirow{3}{*}{0-1 Error} & All &        0.038 &       0.002 &       18.981 &      0.0 &  0.853 &               434 \\
      & Single Model &        0.029 &       0.006 &        5.038 &      0.0 &  0.883 &                54 \\
      & Ensemble &        0.039 &       0.002 &       18.349 &      0.0 &  0.848 &               380 \\
\cline{1-8}
\multirow{3}{*}{NLL} & All &        0.116 &       0.006 &       18.285 &      0.0 &  0.894 &               434 \\
      & Single Model &        0.120 &       0.022 &        5.511 &      0.0 &  0.864 &                54 \\
      & Ensemble &        0.116 &       0.007 &       17.559 &      0.0 &  0.896 &               380 \\
\cline{1-8}
\multirow{3}{*}{Brier} & All &        0.051 &       0.003 &       17.415 &      0.0 &  0.876 &               434 \\
      & Single Model &        0.042 &       0.009 &        4.754 &      0.0 &  0.890 &                54 \\
      & Ensemble &        0.052 &       0.003 &       16.754 &      0.0 &  0.873 &               380 \\
\cline{1-8}
\multirow{3}{*}{rESCE} & All &        0.009 &       0.002 &        4.712 &      0.0 &  0.791 &               434 \\
      & Single Model &        0.026 &       0.007 &        3.755 &      0.0 &  0.632 &                54 \\
      & Ensemble &        0.007 &       0.002 &        3.860 &      0.0 &  0.801 &               380 \\
\bottomrule
\end{tabular}%
}

}
\label{tab:r2_cifar10.1}
\end{table*}

\clearpage 

\subsection{Evaluation on other datasets}
\label{sec:ltrend_other_datasets}
In this section we follow the same conventions as in \cref{fig:ltrend_metrics} to analyze the generalization performance for two other OOD datasets for CIFAR10 and ImageNet, namely CIFAR10.1 and ImageNetC \cite{hendrycks2018benchmarking}. For ImageNetC we focus on our distortions from this dataset; namely brightness, contrast, fog and gaussian noise for three different degrees of corruption. 

\begin{figure*}[!ht]
\centering
\captionsetup{justification=centering}
\subfloat{\includegraphics[width=1\textwidth]{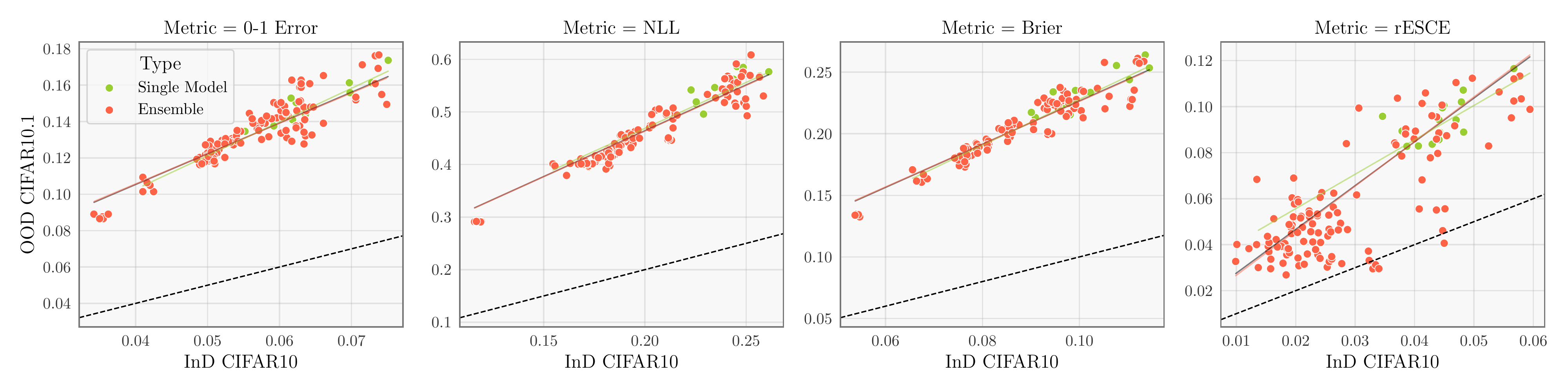}
}
\caption{Generalization Trends for CIFAR10 vs CIFAR10.1.\\ Conventions and conclusions as in \cref{fig:ltrend_metrics}.}
\label{fig:ltrend_cifar10.1}
\end{figure*}

\begin{figure*}[!ht]
\centering
\captionsetup{justification=centering}
\subfloat{\includegraphics[width=1\textwidth]{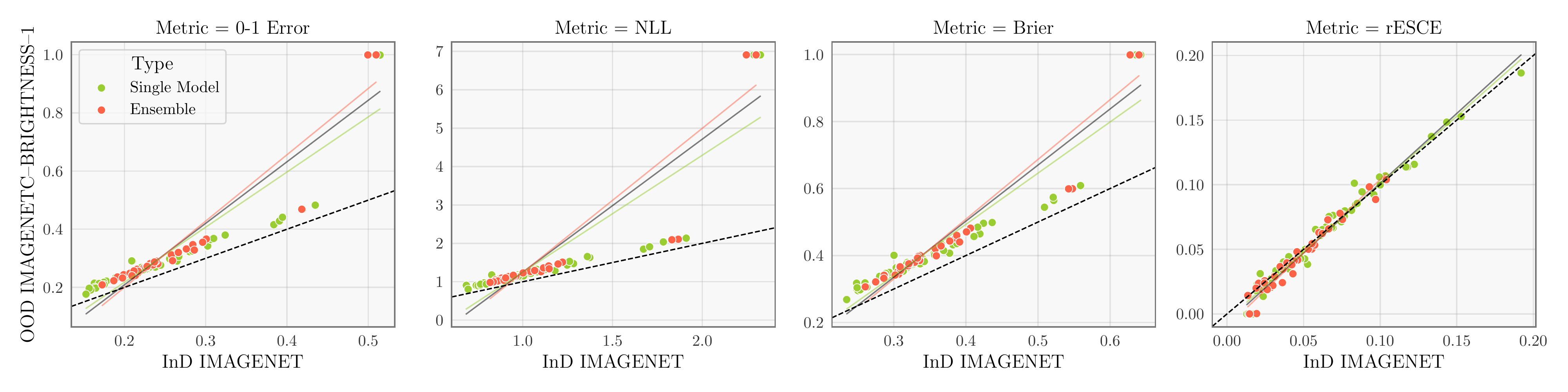}
}
\hfill
\vspace{-0.35in}
\subfloat{\includegraphics[width=1\textwidth]{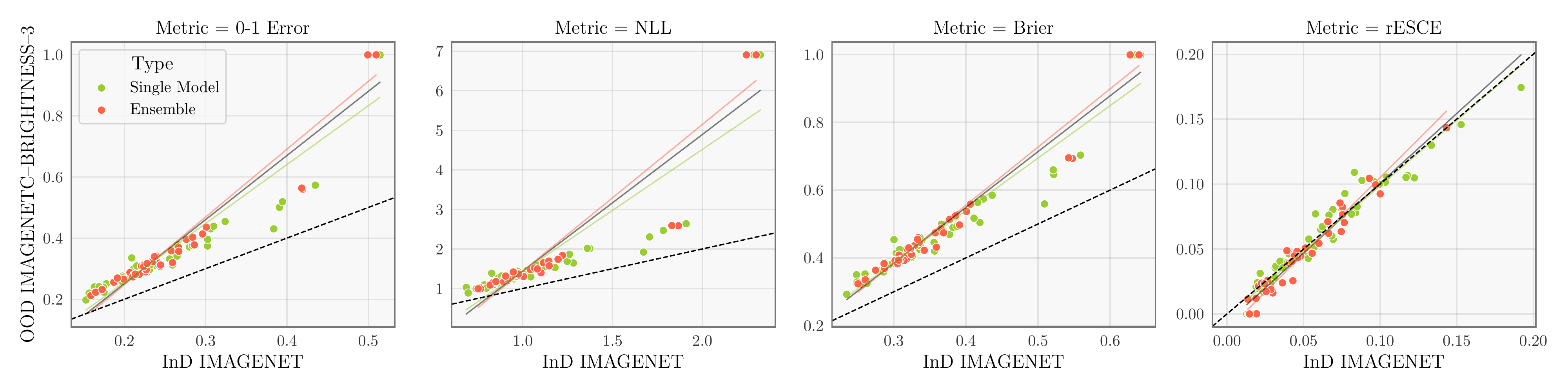}%
}
\hfill
\vspace{-0.2in}
\subfloat{\includegraphics[width=1\textwidth]{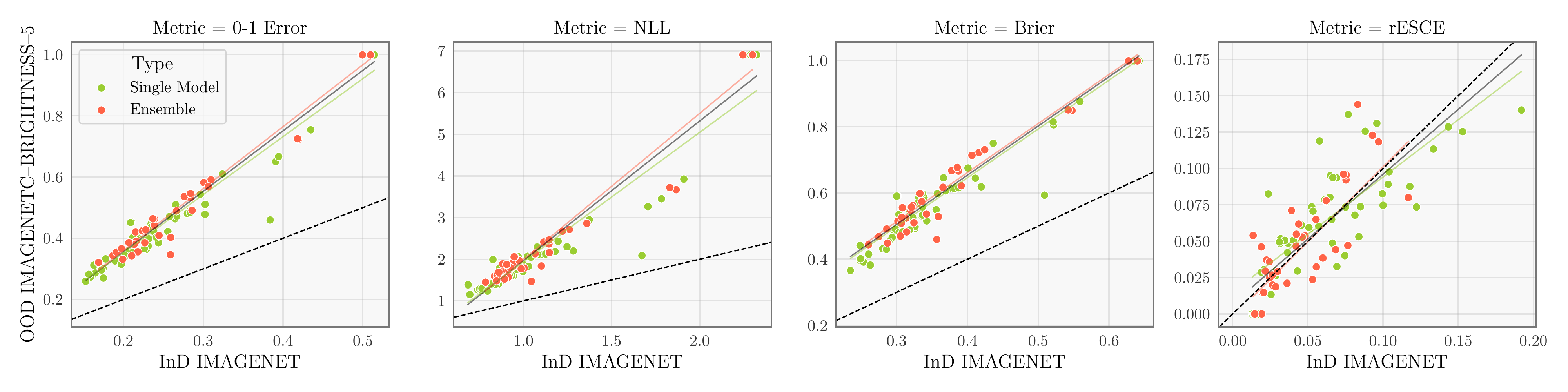}%
}
\caption{Generalization Trends for ImageNet vs ImageNet-C Brightness-1, 3 and 5.\\ Conventions and conclusions as in \cref{fig:ltrend_metrics}.}
\label{fig:imagenetc_brightness1_metrics}
\end{figure*}

\begin{figure*}[!ht]
\centering
\captionsetup{justification=centering}
\subfloat{\includegraphics[width=1\textwidth]{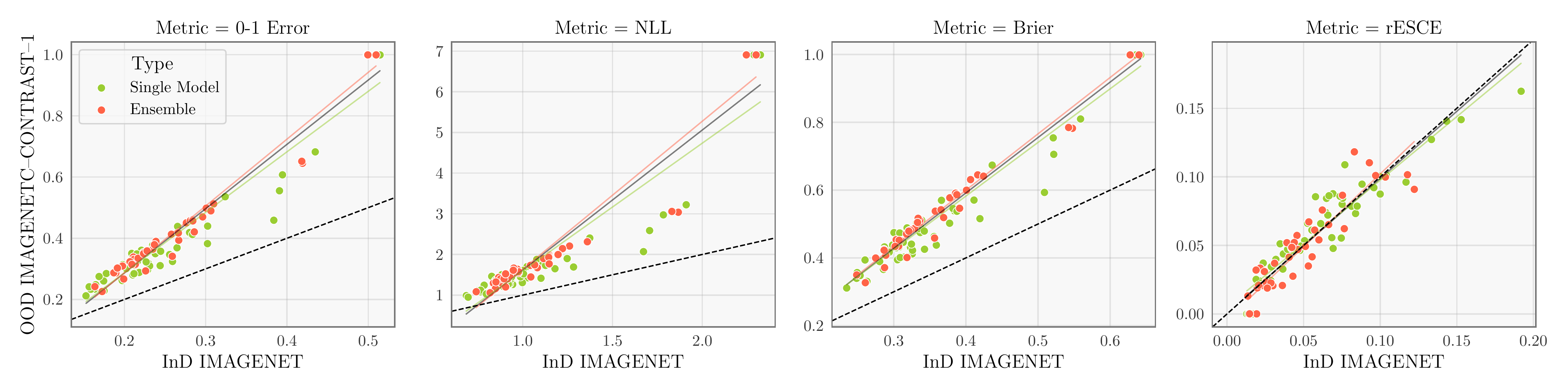}
}
\hfill
\vspace{-0.35in}
\subfloat{\includegraphics[width=1\textwidth]{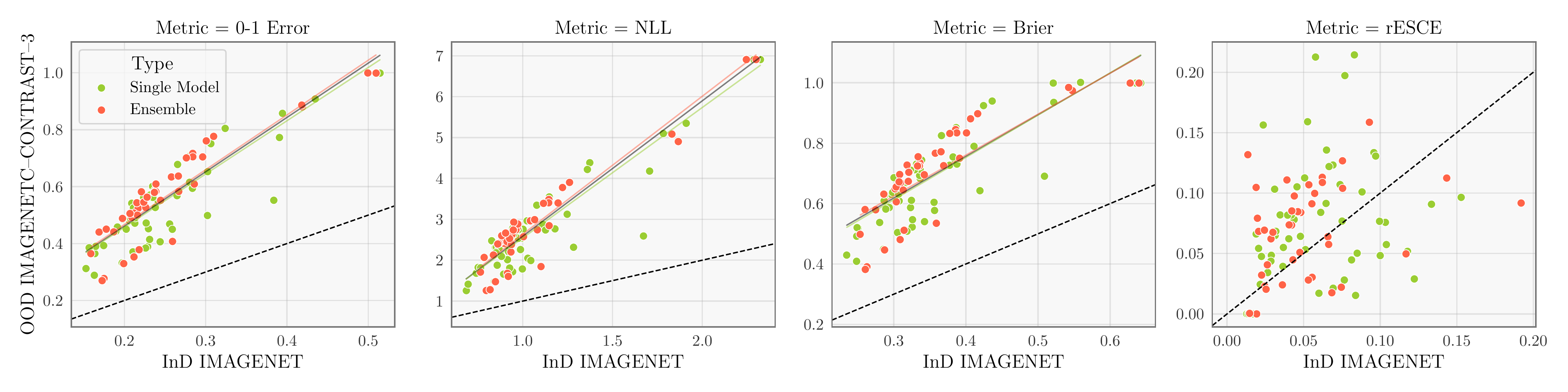}%
}
\hfill
\vspace{-0.2in}
\subfloat{\includegraphics[width=1\textwidth]{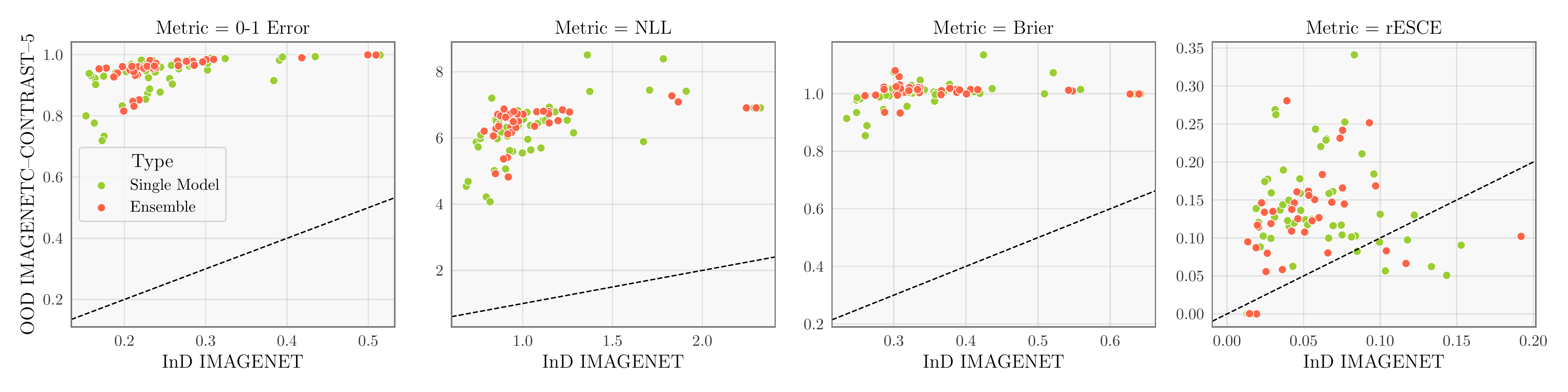}%
}
\caption{Generalization Trends for ImageNet vs ImageNet-C Contrast-1, 3 and 5.\\ Conventions and conclusions as in \cref{fig:ltrend_metrics}.}
\label{fig:imagenetc_contrast1_metrics}
\end{figure*}

\begin{figure*}[!ht]
\centering
\captionsetup{justification=centering}

\subfloat{\includegraphics[width=1\textwidth]{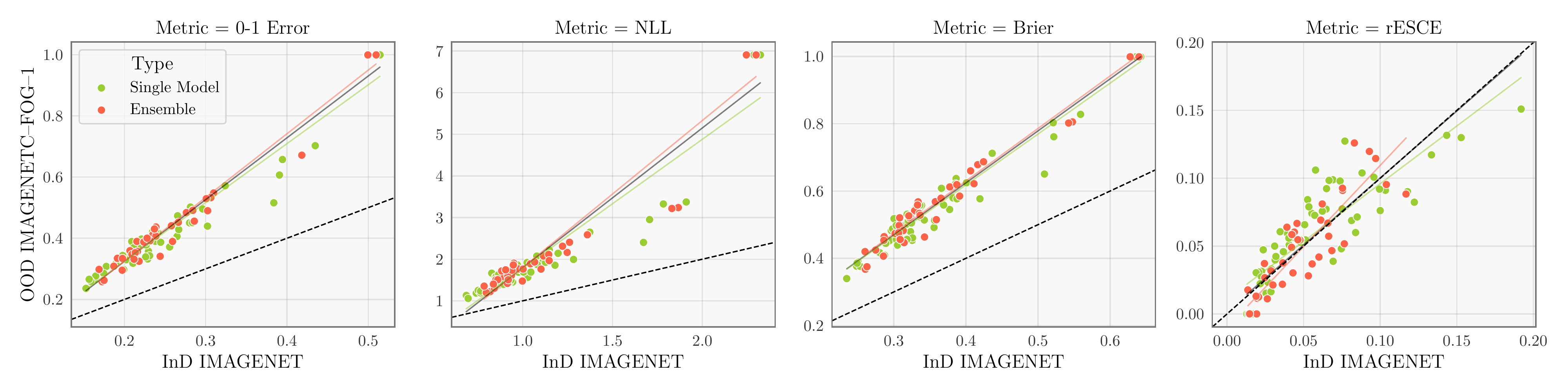}%
}
\hfill
\vspace{-0.35in}
\subfloat{\includegraphics[width=1\textwidth]{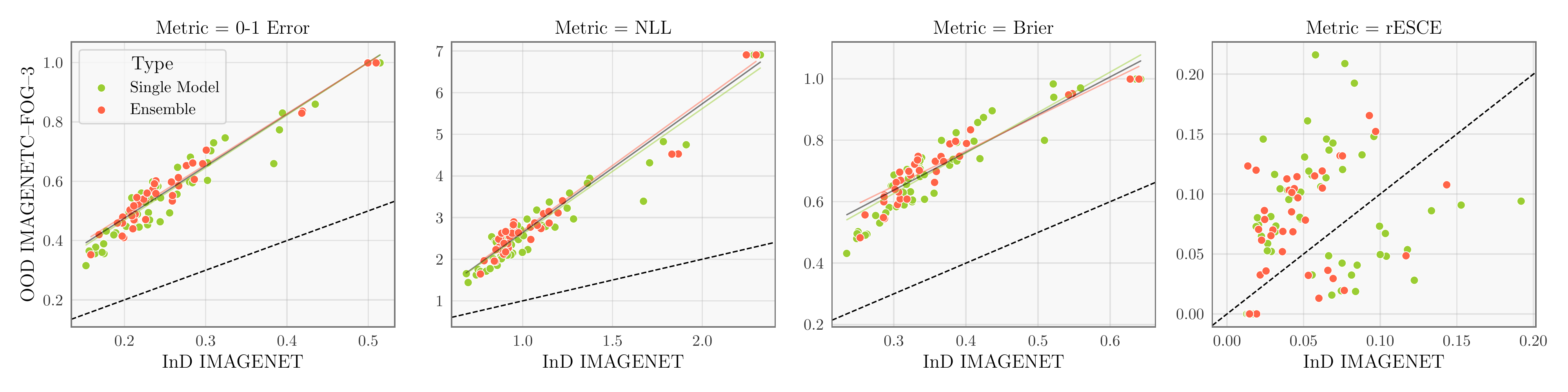}%
}
\hfill
\vspace{-0.2in}
\subfloat{\includegraphics[width=1\textwidth]{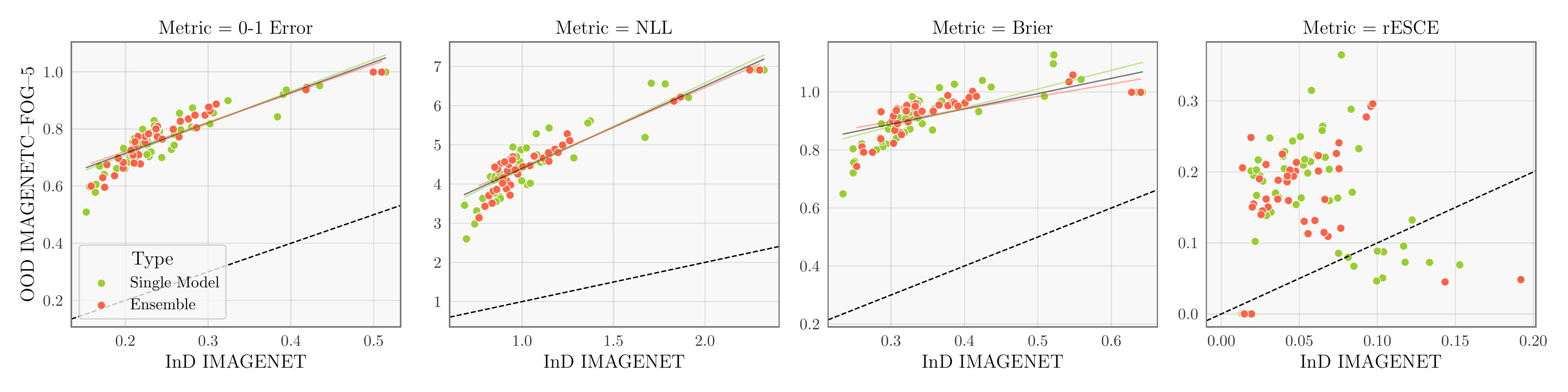}%
}
\caption{Generalization Trends for ImageNet vs ImageNet-C Fog-1,3, and 5.\\ Conventions and conclusions as in \cref{fig:ltrend_metrics}.}
\label{fig:imagenetc_fog1_metrics}
\end{figure*}

\begin{figure*}[!ht]
\centering
\captionsetup{justification=centering}
\subfloat{\includegraphics[width=1\textwidth]{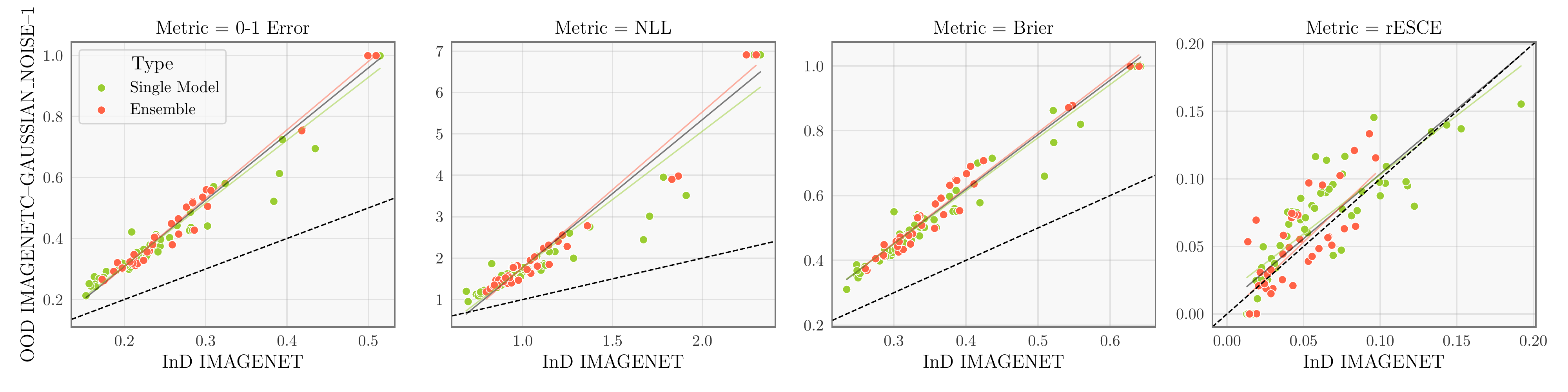}%
}
\hfill
\vspace{-0.35in}
\subfloat{\includegraphics[width=1\textwidth]{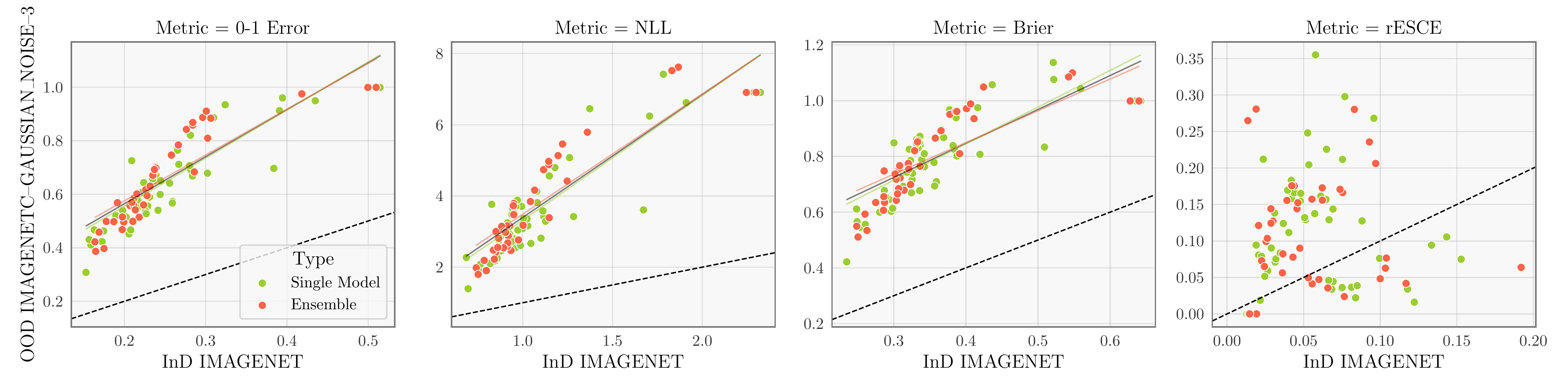}%
}
\hfill
\vspace{-0.2in}
\subfloat{\includegraphics[width=1\textwidth]{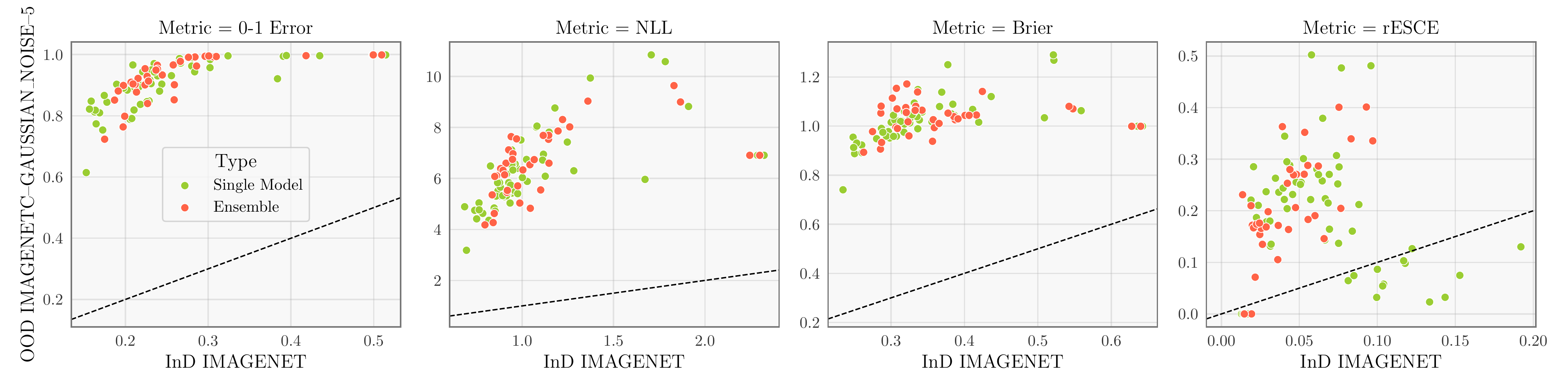}%
}
\caption{Generalization Trends for ImageNet vs ImageNet-C Gaussian Noise-1,3, and 5.\\ Conventions and conclusions as in \cref{fig:ltrend_metrics}.}
\label{fig:imagenetc_gaussiann1_metrics}
\end{figure*}

\clearpage 

\subsection{Calibration metrics}
\label{sec:app_ece_all}

The calibration error is a frequentist idea to measure the quality of uncertainty given by a model. The calibration error is given by,
\begin{align*}
    \text{Calibration Error} = | Confidence - Accuracy |
\label{eq:calibration_error}
\end{align*}

To compare the calibration error across multiple models, practitioners have resorted to the expected calibration error (ECE) \cite{naeini2015obtaining}, which approximates the calibration error by binning the predictive probabilities and taking a weighted average of the calibration errors across bins. ECE provides a scalar summary statistic of the quality of uncertainty \cite{guo2017calibration}.  We employ both the ECE and, a smooth approximation, the root of the Expected Squared Calibration Error (rESCE) to compare the quality of uncertainty gained by ensembling over individual models. The rESCE is defined as,

\begin{align*}
     \text{rESCE} = \sqrt{\sum \frac{\mid B_m\mid}{n}[acc(B_m)-conf(B_m)]^2} 
\end{align*}

\begin{figure*}[h!]
\centering
\captionsetup{justification=centering}
\subfloat{\includegraphics[width=0.9\textwidth]{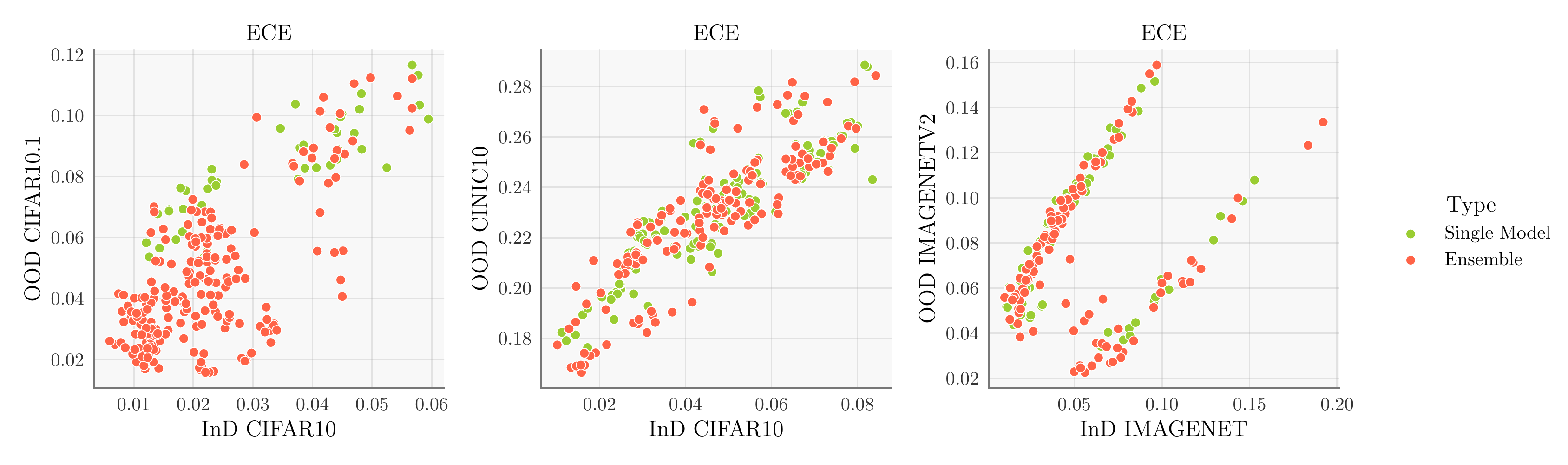}
}
\vspace{-0.1in}
\caption{Generalization trends for the Expected Calibration Error (ECE) for different datasets.\\ Conventions and conclusions similar to the rESCE (right column) in \cref{fig:ltrend_metrics}.}
\label{fig:ece_all}
\end{figure*}

\cref{fig:ltrend_metrics} provides the rESCE for in distribution vs out of distribution for CIFAR10 vs CIFAR10.1 and Imagenet vs ImagenetV2. (See \cref{sec:ltrend_other_datasets} for additional datasets).  In \cref{fig:ece_all}, we evaluate the generalization performance in terms of the ECE, following the conventions in \cref{fig:ltrend_metrics}.
From \cref{fig:ece_all} we see that there is no clear trend for InD versus OOD generalization across different datasets.
Furthermore, we find that---for CIFAR10/CIFAR10.1---ensembles are able to achieve some amount of effective robustness with respect to the ECE metric.
However, for the other two dataset pairs, we find that the ECE performance of ensembles heavily overlaps with the ECE performance of single models for most models.    

\subsection{Comparison between homogeneous, heterogeneous and implicit deep ensembles}
\label{sec:app_heter_ensemble}
\begin{figure*}[h!]
\centering
\captionsetup{justification=centering}
\subfloat{\includegraphics[width=1\textwidth]{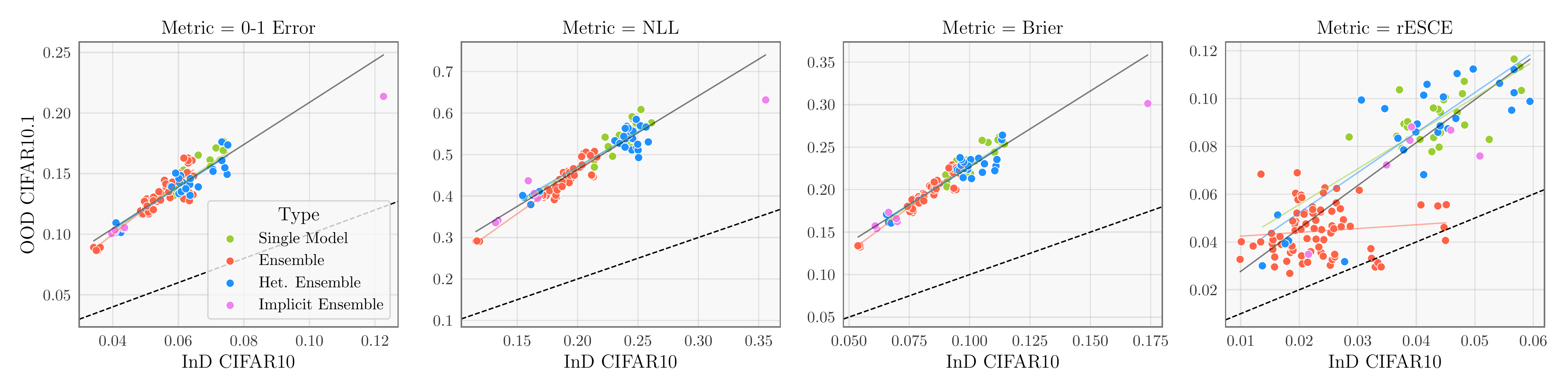}

}
\hfill
\vspace{-0.35in}
\subfloat{\includegraphics[width=1\textwidth]{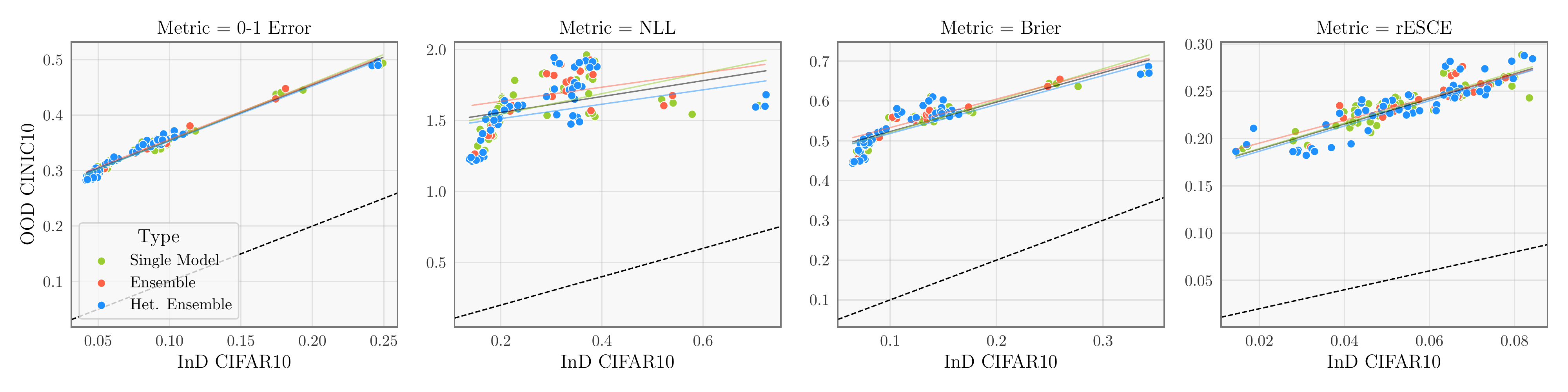}%
}
\hfill
\vspace{-0.2in}
\subfloat{\includegraphics[width=1\textwidth]{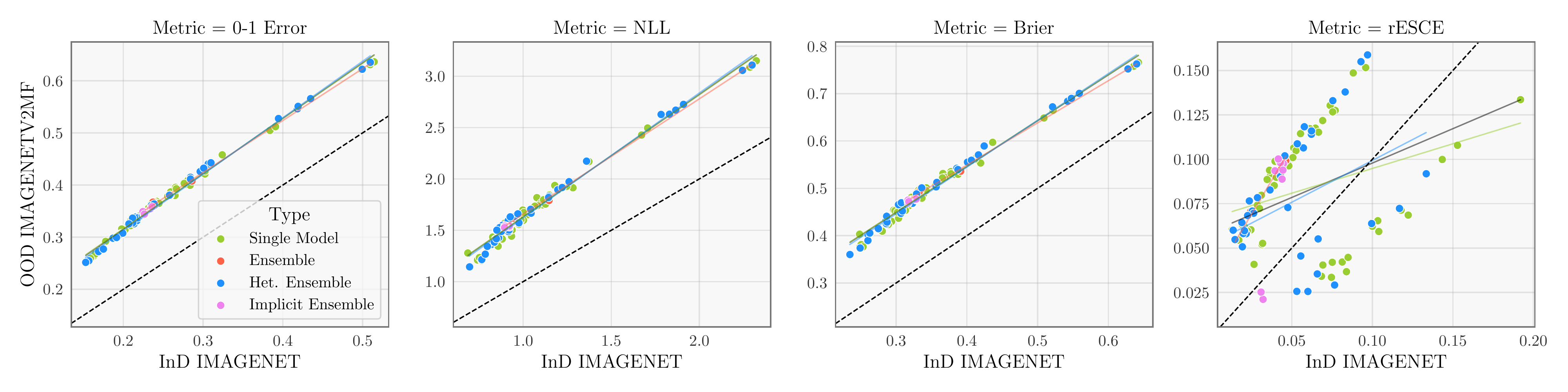}%
}
\caption{Homogeneous and heterogeneous deep ensembles in \cref{fig:ltrend_metrics} follow similar generalization trends.  Conventions and conclusions as in \cref{fig:ltrend_metrics}, including Implicit Deep ensemble models.}
\label{fig:het_ensembles_metrics}
\end{figure*}

In this section we split the data from the ensemble model class in \cref{fig:ltrend_metrics} into two sub classes: an ensemble class which now contains only the homogeneous ensembles, and the heterogeneous class; to explore if the ensemble model classes provide different generalization trends. We find that this is not the case for several in distribution and out of distribution pairs illustrated in \cref{fig:het_ensembles_metrics}.

\ekb{We were additionally interested in evaluating whether implicit deep ensembles, models which aim to bridge the gap between individual networks and deep ensembles such as MC Dropout \cite{gal2016dropout}, Batch Ensemble \cite{wen2020batchensemble}, and MIMO \cite{havasi2021training}, also follow the same observed trends.  We include the performance of implicit ensembles, including MC Dropout and MIMO, in \cref{fig:het_ensembles_metrics}. The implicit deep ensemble models for ImageNet were constructed from a Resnet50 architecture, which was selected given its ubiquitous deployment, and availability in the open source Uncertainty Baselines library \cite{nado2021uncertainty}. The number of implicit models considered for ImageNet is 12 (2 MC dropout models, and 10 for MIMO models). The implicit deep ensemble models for CIFAR10 were constructed from a WideResnet-28 architecture. In total, we considered 6 implicit ensemble models (3 MC dropout models and 3 MIMO models). Overall the results illustrated in \cref{fig:het_ensembles_metrics} show that implicit deep ensembles also fall on the line, along with individual models, heterogeneous, and homogeneous ensembles, and do not constitute an effectively robust model class.}

\begin{figure*}[!ht]
\centering
\captionsetup{justification=centering}
\subfloat{\includegraphics[width=1\textwidth]{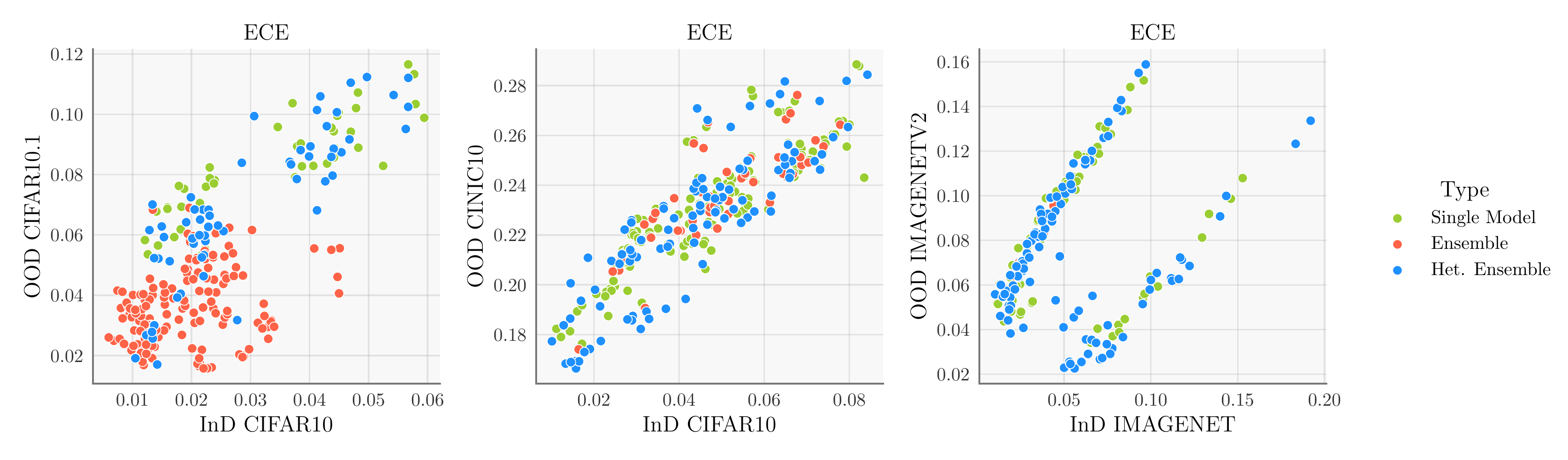}
}
\caption{Expected Calibration Error (ECE) for homogeneous and heterogeneous ensembles follow similar trends as rESCE in \cref{fig:ltrend_metrics}.}
\label{fig:ece_heter}
\end{figure*}

\clearpage 

\clearpage

\newpage

\newpage

\clearpage 

\section{Quantifying improvement similarity between single models and ensembles\label{sec:appmmd}}

In order to validate the statistical significance of the correlations that we observe between the improvements made by ensembles and single models, we considered as baselines the distribution of improvements we would expect from comparing \textit{within} each model type- i.e. improvement correlations comparing the improvement of two performance matched single models, or two performance matched ensembles.
Although Pearson's $R$ provides a good visual aid to interpret the trends visually, we wanted to be more agnostic when validating the trends that we see.
We directly compared the improvement correlations that we see between ensemble/larger model pairs with improvement correlations that resulted when we substituted one member of these pairs with another, similarly performing model from the opposite model type (i.e., replace the ensemble with a control single model, or vice versa). We compared the resulting pair of improvement correlations with a kernel two sample test \cite{gretton2012kernel}. We calculated the unbiased test statistic $MMD_u^{2}$ from this paper for each combination of improvement correlations, and determined an appropriate threshold for these statistics based upon which we would reject the null hypothesis (Corollary 11 in \cite{gretton2012kernel}). 

For each entry in the tables shown here, we consider the performance of four different kinds of predictions: an ensemble, an average single model with similar performance, a ``control" set of average single models or ensembles (again with similar performance) and finally a base model, against which we are comparing the performance of all other models. Each entry compares one of the improvement correlations shown in \cref{fig:perfcomp} (given by the row) against the ``control" improvement correlation listed in the column.  In all comparisons, across InD and OOD data, CINIC10 and CIFAR10.1, and using NLL or Brier Score as metrics, we failed to reject the null hypothesis that the distributions we compared were significantly different. Each table shows the statistic value that we computed from any given pair of improvement correlations, along with the threshold statistic value we would have to exceed to reject the null hypothesis in parentheses. We list details of each comparison with each table. 

Finally, at the end of this section we list the accuracies of the models that we compare, ensuring that the overall performance of these models does not differ too significantly, regardless of the metric under consideration. 

{\renewcommand{\arraystretch}{1.5}
\begin{table*}[h!]
\centering{
\caption{\textbf{ $MMD^2_u$ to compare performance gains: CINIC10 NLL.} Base network: VGG-11. Average Model: WideResNet-18-4. Control Single Model: WideResNet-18-2. Control Ensemble: GoogleNet}
\resizebox{.9\textwidth}{!}{%
\begin{tabular}{l  c  c c c}
\hline
&
& & \multicolumn{1}{l}{$\Delta$ Single/$\Delta$ Ctrl. Single} & \multicolumn{1}{l}{$\Delta$ Ensemble/$\Delta$ Ctrl. Ensemble}   \\ \hline
\multirow{2}{*}{$\Delta$ Ensemble/$\Delta$ Single} 
& CIFAR10 (InD)  & & $2.2\mathrm{e}{-3} (0.069)$ & $2.1\mathrm{e}{-2}  (0.069)$   \\
  & CINIC10 (OOD) &   & $3.3\mathrm{e}{-3}  (0.031)$ & $2.0\mathrm{e}{-3} (0.031)$  \\
\hline
\end{tabular}%
}
}
\label{tab:MMD1}
\end{table*}
}

{\renewcommand{\arraystretch}{1.5}
\begin{table*}[h!]
\centering{
\caption{\textbf{ $MMD^2_u$ to compare performance gains: CINIC10 Brier Score}. Base network: VGG-11. Average Model: WideResNet-18-4. Control Single Model: WideResNet-18-2. Control Ensemble: GoogleNet }
\resizebox{.9\textwidth}{!}{%
\begin{tabular}{l  c  c c c}
\hline
&
& & \multicolumn{1}{l}{$\Delta$ Single/$\Delta$ Ctrl. Single} & \multicolumn{1}{l}{$\Delta$ Ensemble/$\Delta$ Ctrl. Ensemble}   \\ \hline
\multirow{2}{*}{$\Delta$ Ensemble/$\Delta$ Single} 
& CIFAR10 (InD)  & & $2.7\mathrm{e}{-4} (0.069)$ & $1.3\mathrm{e}{-2}  (0.069)$   \\
  & CINIC10 (OOD) &   & $4.2\mathrm{e}{-3}  (0.031)$ & $4.8\mathrm{e}{-3} (0.031)$  \\
\hline
\end{tabular}%
}
}
\label{tab:MMD2}
\end{table*}
}

{\renewcommand{\arraystretch}{1.5}
\begin{table*}[h!]
\centering{
\caption{\textbf{ $MMD^2_u$ to compare performance gains: CIFAR10.1 NLL}. Base network: ResNet 18. Average Model: WideResNet 18-4. Control Single Model: WideResNet 18. Control Ensemble: VGG 11 }
\resizebox{.9\textwidth}{!}{%
\begin{tabular}{l  c  c c c}
\hline
&
& & \multicolumn{1}{l}{$\Delta$ Single/$\Delta$ Ctrl. Single} & \multicolumn{1}{l}{$\Delta$ Ensemble/$\Delta$ Ctrl. Ensemble}   \\ \hline
\multirow{2}{*}{$\Delta$ Ensemble/$\Delta$ Single} 
& CIFAR10 (InD)  & & $2.4\mathrm{e}{-2} (0.069)$ & $5.1\mathrm{e}{-3}  (0.069)$   \\
  & CIFAR10.1 (OOD) &   & $8.1\mathrm{e}{-3}  (0.15)$ & $2.0\mathrm{e}{-3} (0.15)$  \\
\hline
\end{tabular}%
}
}
\label{tab:MMD3}
\end{table*}
}

{\renewcommand{\arraystretch}{1.5}
\begin{table*}[h!]
\centering{
\caption{\textbf{ $MMD^2_u$ to compare performance gains: CIFAR10.1 Brier Score}. Base network: ResNet 18. Average Model: WideResNet 18-4. Control Single Model: WideResNet 18. Control Ensemble: VGG 11}
\resizebox{.9\textwidth}{!}{%
\begin{tabular}{l  c  c c c}
\hline
&
& & \multicolumn{1}{l}{$\Delta$ Single/$\Delta$ Ctrl. Single} & \multicolumn{1}{l}{$\Delta$ Ensemble/$\Delta$ Ctrl. Ensemble}   \\ \hline
\multirow{2}{*}{$\Delta$ Ensemble/$\Delta$ Single} 
& CIFAR10 (InD)  & & $2.1\mathrm{e}{-2} (0.069)$ & $2.5\mathrm{e}{-3}  (0.069)$   \\
  & CIFAR10.1 (OOD) &   & $8.1\mathrm{e}{-3}  (0.15)$ & $8.7\mathrm{e}{-4} (0.15)$  \\
\hline
\end{tabular}%
}
}
\label{tab:MMD4}
\end{table*}
}

{\renewcommand{\arraystretch}{1.5}
\begin{table*}[h!]
\centering{
\caption{\textbf{Corresponding accuracies for models compared on CIFAR10/CINIC10. } Architectures of models left to right: VGG-11, WideResNet-18-4, WideResNet-18-2, GoogleNet}
\resizebox{1.0\textwidth}{!}{%
\begin{tabular}{l c c c c c c}
\hline
Dataset & Ensemble & Single Model & Single Model Control & Ensemble Control \\ \hline

CIFAR10 (InD)   & $93.44\%$ & $94.32\pm0.1495\%$ & $93.93\pm0.0804\%$ & $93.68\%$ & \\
CINIC10 (OOD)   & $67.88\%$ & $68.84\pm0.3936\%$ & $68.59\pm0.2762\%$ & $69.10\%$ &  \\
\hline
\end{tabular}%
}
}
\label{tab:MMD5}
\end{table*}
}

{\renewcommand{\arraystretch}{1.5}
\begin{table*}[h!]
\centering{
\caption{\textbf{Corresponding accuracies for models compared on CIFAR10/CIFAR10.1. } Architectures of models left to right: ResNet18, WideResNet18-4, WideResNet18, VGG-11}
\resizebox{1.0\textwidth}{!}{%
\begin{tabular}{l c c c c c c }
\hline
Dataset & Ensemble & Single Model & Single Model Control & Ensemble Control \\ \hline

CIFAR10 (InD)   & $94.26\%$ & $94.28\pm0.1430\%$ & $92.77\pm0.1839\%$ & $93.72\%$ &  \\
CIFAR10.1 (OOD) & $86.10\%$ & $86.26\pm0.1727\%$ & $84.61\pm0.4242\%$ & $84.12\%$ &   \\
\hline
\end{tabular}%
}
}
\label{tab:MMD6}
\end{table*}
}

\end{document}